\documentclass[letterpaper]{article}
\usepackage{aaai}
\usepackage{times}
\usepackage{helvet}
\usepackage{courier}
\usepackage{booktabs}
\usepackage{amsmath}
\usepackage{graphicx}
\usepackage{xcolor}
\usepackage[hidelinks]{hyperref}
\ifdefined
\fi
\begin{document}

\title{Ko-WideSearch: A Korean Breadth-Search Benchmark for\\Exhaustive Set Enumeration by Web Agents}
\author{Minbyul Jeong \\
Upstage AI \\
minstar@upstage.ai
}
\maketitle

\begin{abstract}
\begin{quote}
Web-agent benchmarks overwhelmingly measure depth---pinning one obscure answer behind a chain of
constraints---while breadth, exhaustively enumerating a closed set and filling each item's attributes,
is barely evaluated, especially outside English. Breadth is also hard to build: certifying that a gold
set is complete and every cell correct is far costlier than checking a single answer. I introduce
\textsc{Ko-WideSearch}, a Korean breadth-search benchmark built by an automated synthesize-and-verify
pipeline. Each task names a set-parent entity---a TV season, a dynasty, a league, an administrative
region, an election---and asks for its full membership plus a per-item attribute table, graded by Item-,
Column-, and Row-F1. It spans 228 tables over 190 entities and sixteen categories across three difficulty
tiers, set by two structural knobs I dial independently---table width and a 2-D composite key---so
cross-product membership climbs from 0\% to 100\% across the tiers. A single normalization-aware
comparator is shared between gold construction and grading, so stable date and count columns are not
over-dropped on formatting alone. Across twenty web agents, the failure is consistent: agents recover the
set but not the rows (e.g.\ Item-F1 92.8 against Row-F1 53.7), accuracy falls steadily as the knobs
harden, and neither more search nor more spend closes the gap. Broken down by cell, the hard part is
finding the right value, not formatting it: open-ended free-text cells fail most, while cells with a
standard answer such as a date or a name usually come out right.
\end{quote}
\end{abstract}

\section{Introduction}

\begin{figure*}[t]
\centering
\includegraphics[width=\textwidth]{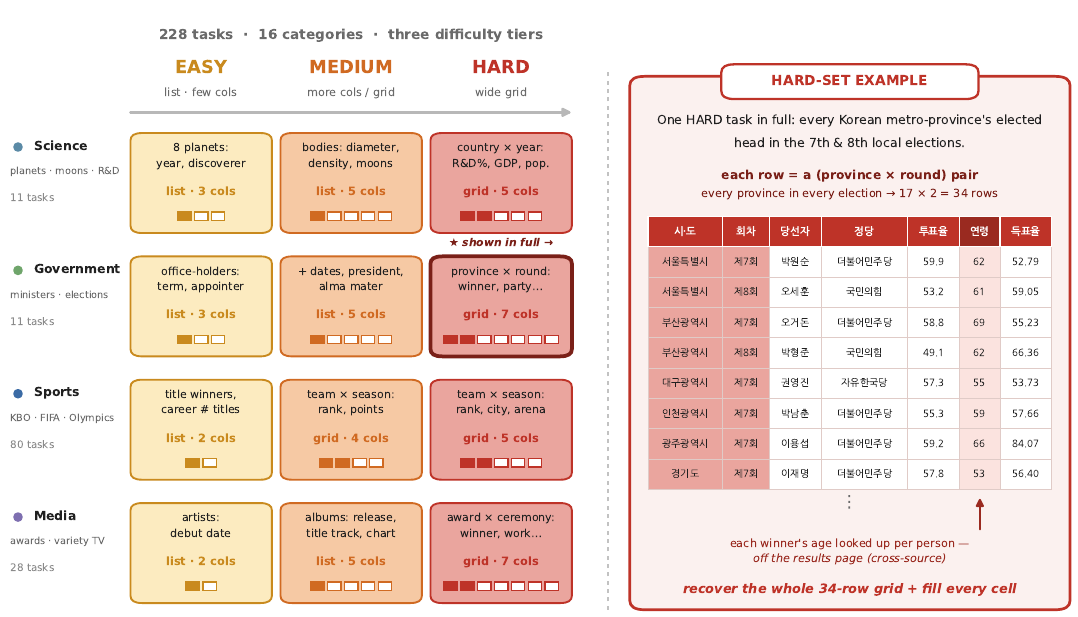}
\caption{\textbf{The benchmark and its running example.} \textbf{Left:} tasks span Korean categories (rows)
and three difficulty tiers (columns); each cell is one real task, with a strip of its columns
beneath---the membership-key column(s) shaded, one box marking a list, two a cross-product
grid. Reading left to right, tasks gain columns to fill and their membership turns from a list
into a grid. \textbf{Right:} the Government/\textsc{Hard} cell shown in full---a \textbf{hard-set
example} I reuse as the paper's running example: every Korean metropolitan province's elected head in
the 7th and 8th local elections. Its rows are (province\,$\times$\,round) pairs (every province in every election,
$17\times2=34$ rows), and per-row attributes such as each winner's age are looked up off the results
page (cross-source). Four of sixteen categories are shown.}
\label{fig:motivation}
\end{figure*}

Ask a web agent to list every low-cost carrier flying in Korea as of June 2026, and for each
one its parent company, the year it launched, its main hub, and how many aircraft it operates today.
A capable agent has to do two things at once. 
It must recover the complete set---nine carriers, where dropping Aero\,K or inventing a tenth is immediately wrong---;
and it must fill, for each row, a cell that lives on a different page, because the current fleet count is not printed beside the airline's name.
Neither half is a hard single-fact lookup. The difficulty is that the
answer is a whole table, and it has to be exhaustive and internally complete.

This is \emph{breadth}---a different axis from the one today's browsing-agent benchmarks
measure. The dominant paradigm is depth: BrowseComp~\cite{wei2025browsecomp} and its Chinese and Korean
siblings BrowseComp-ZH~\cite{zhou2025browsecompzh} and K-BrowseComp~\cite{lee2026kbrowsecomp} hide a
single short answer behind several constraints, and the agent traverses a multi-hop or
parallel-constraint path to recover it. WideSearch~\cite{widesearch2025} showed that even
frontier agents fail set-valued tasks, not because the information is hidden but because of
scale and exhaustiveness: they drop rows, mis-fill cells, and lose track of the set boundary.
Breadth, in other words, stresses a capability that depth benchmarks never exercise---holding and
completing a structured set across many pages.

Two gaps motivate this work. First, breadth-search evaluation barely exists outside English: Korean
agentic benchmarks are scarce~\cite{lee2026kbrowsecomp}, yet browsing ability is uniquely language- and
culture-bound---an agent must navigate Korean sources whose structure, terminology, and search
conventions differ from English ones~\cite{kim2024click,son2025kmmlu}. Second, breadth poses a
construction problem of its own: certifying that a gold set is complete and that every attribute
cell is correct is far harder than checking one answer, and doing it at scale by hand is
prohibitive---the original WideSearch was hand-built at 200 tables and high annotation cost.

I address both with \textsc{Ko-WideSearch}, a Korean breadth-search benchmark built by an automated
synthesize-and-verify pipeline. Each task names a set-parent entity---a TV season, a dynasty, a
league, an administrative region, an election---and asks for the full membership plus an attribute
table, graded by Item-, Column-, and Row-F1. I organize the benchmark---228 tables in all---into
three difficulty tiers along two structural axes I dial independently: table width and a
2-D composite key (Figure~\ref{fig:motivation}).
My contributions are fourfold. (1)~I introduce \textsc{Ko-WideSearch}: 228 Korean breadth-search
tables---4{,}262 gold rows and 14{,}560 attribute cells over 190 set-parent entities and sixteen
categories---every table web-sourced and decontaminated. (2)~I build it with an automated synthesize-and-verify pipeline
that makes Korean breadth-search evaluation scalable while keeping the gold trustworthy: a build agent
enumerates each gold table by exhaustive web search, three independent gates certify non-memorizability,
completeness, and cross-source attribute verification, and two difficulty knobs manufacture
WideSearch-grade hard breadth---2-D membership rises from 0\% in \textsc{Easy} to 100\% in \textsc{Hard},
and median width from three to seven columns. (3)~I find that a na\"ive attribute cross-check over-drops
stable columns on formatting alone and, because the same comparator grades model output, would otherwise
mis-score agents, so I fix both with one normalization-aware comparator; separately, I web-ground the
single-page-versus-cross-source sourcing label, which an LLM guesses only roughly 72\% accurately against
the live web. (4)~I adopt a leakage-aware release: because an agent's own search can surface posted gold
and lift the answer, I follow the held-out precedent of GAIA~\cite{mialon2023gaia} and
BrowseComp~\cite{wei2025browsecomp} and open-source the pipeline and scorer under MIT while distributing
the evaluation data by request, keeping the set off the surface agents search so Korean breadth search
can be evaluated and re-grown.

\section{Related Work}
\paragraph{Web agents, tool use, and browsing benchmarks.}
A browsing agent is a language model that acts---reasoning, calling tools, and reading back what it
finds---a loop established by ReAct~\cite{yao2023react} and by tool-augmented models that teach
themselves to invoke APIs~\cite{schick2023toolformer,patil2024gorilla,qin2024toolllm}. Pointed at the
open web, such agents have been studied as browser-assisted question answerers~\cite{nakano2021webgpt}
and as navigators of real and simulated
sites~\cite{yao2022webshop,deng2023mind2web,he2024webvoyager,zhou2024webarena}, and stress-tested by
agent suites such as AgentBench~\cite{liu2024agentbench}, GAIA~\cite{mialon2023gaia}, and
AssistantBench~\cite{yoran2024assistantbench}; more recently, search has been folded into
reasoning itself~\cite{li2025searcho1}. The hardest browsing benchmarks measure depth:
BrowseComp~\cite{wei2025browsecomp} poses questions a human cannot answer quickly with a browser,
solvable only through multi-step search, and BrowseComp-ZH~\cite{zhou2025browsecompzh} and
K-BrowseComp~\cite{lee2026kbrowsecomp} carry that paradigm to Chinese and Korean---the latter
emphasizing Korean-context grounding (local entities, semi-structured Korean pages, culturally grounded
clues), whose construction-and-verification methodology I build on. This depth lineage runs back
through multi-hop and open-domain QA---HotpotQA~\cite{yang2018hotpotqa},
TriviaQA~\cite{joshi2017triviaqa}, Natural Questions~\cite{kwiatkowski2019nq},
2WikiMultihopQA~\cite{ho2020twowiki}, and MuSiQue~\cite{trivedi2022musique}---and into short-answer
factuality and freshness~\cite{krishna2024frames,vu2024freshllms,wei2024simpleqa,phan2025hle}. Every
one of these, however, grades a single answer per question; \textsc{Ko-WideSearch} grades a
closed set.

\paragraph{Breadth, set enumeration, and tables.}
WideSearch~\cite{widesearch2025} introduced set-valued enumeration as a distinct agentic challenge,
with item-, column-, and row-level F1 as its metrics and high human agreement on gold construction;
follow-up work isolates aggregation and counting error~\cite{aggbench2026}. The set-valued setting has
older roots: in list-answer question answering, where one query has many correct answers scattered
across pages~\cite{amouyal2023qampari}, and in querying semi-structured tables---compositional question
answering over Wikipedia tables~\cite{pasupat2015wtq}, table-grounded fact
verification~\cite{chen2020tabfact}, free-form table QA~\cite{nan2022fetaqa}, and questions that hop
between a table and its linked text~\cite{chen2020hybridqa}. What breadth adds is that the agent must
produce the table rather than read a given one, assembling membership and attributes from the
live web---a retrieval-grounded generation problem in the lineage of
RAG~\cite{lewis2020rag,asai2024selfrag}. I adopt WideSearch's task shape and metrics, instantiate them
on Korean sources, replace its hand construction with an automated, verified pipeline, and add two
structural difficulty knobs (width and a 2-D composite key) that let the pipeline reach the hand-built
original's width and cross-product share, exceeding both in the \textsc{Hard} tier. To my knowledge,
breadth search on Korean sources has no prior benchmark.

\paragraph{Korean and regional evaluation.}
Korean evaluation has matured along a static axis: reading comprehension in
KorQuAD~\cite{lim2019korquad}, core understanding in KLUE~\cite{park2021klue} and
KoBEST~\cite{jang2022kobest}, multitask knowledge in KMMLU~\cite{son2025kmmlu}, factual and cultural
knowledge in HAE-RAE~\cite{son2024haerae} and CLIcK~\cite{kim2024click}, and social bias in
KoBBQ~\cite{jin2024kobbq}; multilingual efforts such as MENLO~\cite{whitehouse2026menlo} and
INCLUDE~\cite{romanou2025include} add regionally grounded knowledge. A fast-growing set of Korean and
Korean-capable models has appeared in parallel---HyperCLOVA~\cite{kim2021hyperclova} and
HyperCLOVA~X~\cite{yoo2024hyperclovax}, EXAONE~\cite{lgai2024exaone}, Kanana~\cite{kakao2025kanana}, and
SOLAR~\cite{kim2024solar}---and I evaluate a Korean-specialized family on the breadth-search task. Valuable as they are, these
benchmarks are largely static: they do not require an agent to search the live web, maintain evidence
state, or synthesize information across Korean pages---exactly the gap a breadth-search benchmark fills.

\begin{figure*}[t]
\centering
\includegraphics[width=\textwidth]{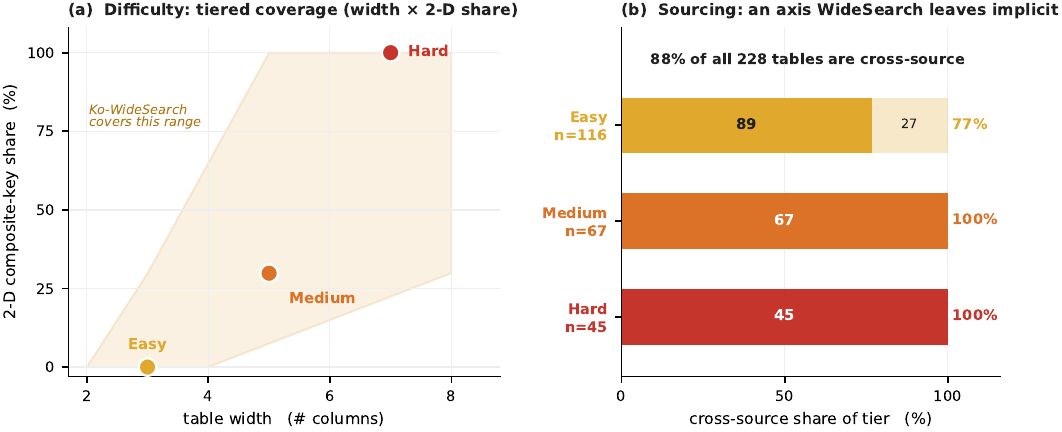}
\caption[Ko-WideSearch extends WideSearch along two structural axes.]{\textbf{\textsc{Ko-WideSearch} extends WideSearch}~\cite{widesearch2025} along two structural
axes. \textbf{(a) Difficulty}: the two knobs I dial---table width and the 2-D composite-key
share---span the plane. WideSearch is a single, un-tiered reference (median six columns, 35\% 2-D),
whereas \textsc{Ko-WideSearch} covers a calibrated
\textsc{Easy}$\rightarrow$\textsc{Medium}$\rightarrow$\textsc{Hard} region (shaded), its
\textsc{Hard} tier reaching the original's width and exceeding its cross-product share (100\% vs.\ 35\%).
\textbf{(b) Sourcing}: an orthogonal, web-grounded property---whether a table's attributes sit on one
page (\textsc{exhaustive}) or span several (\textsc{cross-source})---rising from 77\% cross-source in
\textsc{Easy} to every \textsc{Medium} and \textsc{Hard} task.}
\label{fig:composition}
\vspace{-0.2cm}
\end{figure*}

\paragraph{Synthetic data, judging, and contamination.}
My pipeline sits in the lineage of scaling data by self-generation, from bootstrapping instructions out
of a model's own outputs~\cite{wang2023selfinstruct,xu2024wizardlm} to synthesizing instruction and
conversation data at scale~\cite{ding2023ultrachat,mukherjee2023orca,xu2024magpie}. But it inverts the
usual goal: I synthesize an evaluation set, where correctness, not diversity, is the binding
constraint. Two risks follow. First, model-generated gold needs trustworthy verification; my acceptance
gates use held-out LLM judges, a now-standard instrument whose human agreement and biases are
documented~\cite{liu2023geval,zheng2023mtbench}. Second, model-generated tasks can be under-specified,
too easy, or already seen in pretraining, so contamination is a central quality
concern~\cite{dong2024generalization,spiesberger2026softcontamination} with a matching line of
detection methods~\cite{sainz2023contamination,golchin2024timetravel,oren2024proving}. I answer both
with verifiability-first construction---every gold set is independently re-enumerated and
search-certified---and an explicit decontamination screen against existing evaluation sets.

\section{The Ko-WideSearch Benchmark}
I build \textsc{Ko-WideSearch} in five steps, taken up in turn below. I first fix the task and its
four metrics. An autonomous build-and-verify pipeline then constructs each gold table and certifies it
behind three independent gates. Two structural knobs---table width and a 2-D composite
key---set the three difficulty tiers. A single normalization-aware comparator, shared between gold
construction and grading, keeps stable columns from being mistaken for source-fragile ones. Finally,
because an LLM cannot reliably guess whether a table's content sits on one page or spans many, I
web-ground an orthogonal sourcing label.

\subsection{Task Definition}
\label{sec:task}
A \textsc{Ko-WideSearch} instance is a question that (i) names a closed, finite set through a predicate
$Y$ and (ii) requests $m{-}k$ attributes for each member. The gold answer is a set of $n$ rows, each a
(name, attributes) pair over $m$ columns; the first $k$ columns are the membership key ($k{=}1$
for an ordinary primary key, $k{=}2$ for a 2-D cross-product such as team\,$\times$\,season) and the
rest are attributes. The low-cost-carrier task above is a small instance: nine rows, keyed on the
carrier name ($k{=}1$), plus four attribute columns (parent company, launch year, hub, fleet size)
for five in all. The
hard-set example I trace through the paper (Figure~\ref{fig:motivation}) is the $k{=}2$ case:
its rows are keyed on a (province, election round) pair, so membership is itself a
grid---$17\times2=34$ rows. A model is
graded by parsing its predicted table and matching it to the gold on the key columns:
\begin{itemize}
\item \textbf{Item-F1}: set-membership precision/recall over the row keys.
\item \textbf{Column-F1}: per-attribute cell correctness over matched rows (micro- and
macro-averaged).
\item \textbf{Row-F1}: the fraction of rows whose key and every attribute cell are
correct---the strict, end-to-end metric.
\item \textbf{Table success}: the fraction of tables that are entirely correct (Row-F1 $=1$),
matching WideSearch's headline pass criterion.
\end{itemize}
Cells are compared with a shared type-aware comparator (below); genuinely-absent gold cells are marked
with a sentinel and excluded from scoring. Each column carries a WideSearch-style metric
declaration---exact match, numeric or date tolerance, URL match, or, for free-text values, an LLM-judge
criterion---but the released numbers are produced by the deterministic comparator throughout, which
resolves a free-text cell by a conservative normalized-text match rather than by calling a judge.

\subsection{Construction Pipeline}
\label{sec:pipeline}
Each table is produced by an autonomous pipeline and then certified by independent gates.

\paragraph{Build.} A build agent receives a set-parent seed entity and, over the
\texttt{search}/\texttt{open}/\texttt{find} tool namespace, designs a bounded enumeration question and
constructs the gold table by exhaustive search. Time-volatile attributes are pinned to an explicit
``as-of'' date (e.g.\ 2026-06) so the answer is stable; time-invariant tables carry no such date.

\paragraph{Verify.} Acceptance requires passing, independently:
\begin{enumerate}
\item \textbf{Non-memorizability.} A closed-book model must not reproduce the gold cells from
memory. I score item and attribute cells, not just names, so a set whose members are
well-known but whose attributes require lookup is still non-memorizable. The gate fails closed:
an inconclusive check is treated as a rejection.
\item \textbf{Completeness.} An independent agent re-enumerates the membership from the question
alone; the two sets must agree (set-F1 $\geq\tau$), my proxy that the gold set is complete and the
boundary unambiguous.
\item \textbf{Cross-source attribute verification.} A separate pass re-looks-up each attribute for the
known members; a column whose values do not agree with this independent source is dropped as
source-fragile. Acceptance requires at least one cross-verified attribute column to survive: a
name-only list is not a valid breadth task.
\end{enumerate}

\subsection{Difficulty Tiers}
\label{sec:tiers}
Breadth has two structural sources of hardness, and I dial them separately. The first is
width: how many attributes per row must be filled, each potentially from its own page. The
second is a 2-D composite key: when membership is itself a cross-product---every (team, season)
or every (election, candidate) pair---the agent must enumerate a grid rather than a list, and a single
missed combination breaks an entire band of rows. The carrier task turns only the first
knob---wide (five columns) but one-dimensional. My running hard-set example
(Figure~\ref{fig:motivation}) turns both: ``for every metropolitan province, name the head
elected in the 7th and 8th local elections, with party, turnout, the winner's age, and vote share.''
Membership is now a (province\,$\times$\,round) grid---seventeen provinces times two elections,
thirty-four rows---so a province missed in either round breaks an entire band, and the winner's age,
off on a different page from the result, makes the seven columns cross-source as well.

These two knobs define three tiers (Table~\ref{tab:stats}, Figure~\ref{fig:motivation}). \textsc{Easy} turns neither: a narrow 1-D
list, a median of three columns and no composite key. \textsc{Medium} turns exactly one---either a
wide 1-D table (many per-item attributes, a median of five columns) or a narrow
2-D grid---so 2-D membership appears in 30\% of \textsc{Medium} tables. \textsc{Hard} turns both: a wide
2-D cross-product grid, a median of seven columns and a composite key in 100\% of tables. Difficulty
thus shifts from row count toward width and grid structure as the tier hardens. Crucially, the
difficulty is not a labeling artifact
but a generation target: the build prompt and acceptance gates were extended to manufacture wide
and 2-D tables, which is how an automated Korean pipeline reaches the column-count and cross-product
proportions of the hand-built original, exceeding them in the \textsc{Hard} tier
(Figure~\ref{fig:composition}(a) positions my three tiers against WideSearch).

\subsection{Normalization-Aware Cell Comparison}
\label{sec:cells}
A subtlety I found to matter in practice: the attribute cross-check above compares two
independently-retrieved values, and is only as good as its equality test. A na\"ive comparator
over-drops stable columns---presidential terms (``1948--1960'' vs.\ a full ISO date range),
heritage-designation dates, or planetary diameters (``12,742\,km'')---because of date-granularity,
thousands-separator, and unit mismatches rather than genuine source-fragility: a formatting
failure dressed up as fragility. The same comparator also grades model answers, so
the bug would also mis-score agents---marking a correct ``12,742\,km'' wrong against a gold
``12742''. I therefore use a single type-aware comparator, shared between gold construction and
grading, that (i) compares dates at their common granularity (a year-only value matches a full date
with the same year), (ii) strips thousands separators and trailing units before numeric comparison and
applies a relative tolerance for large measurements (with AGGBench-style tolerance for
counts~\cite{aggbench2026}), and (iii) falls back to normalized text matching with substring and
token-overlap for names and locations. With this comparator the cross-check keeps the stable
columns and drops only the genuinely fragile ones, such as volatile rankings or ambiguous role labels.

\subsection{Web-Grounded Sourcing Tiers}
\label{sec:grounding}
Orthogonal to the difficulty tiers is a table's sourcing---whether its content sits on one
page or spans several. An
\textsc{exhaustive-only} table has its entire content---membership and all attributes---consolidated on
a single page, so the difficulty is breadth alone; a \textsc{cross-source} table has at least one
attribute that is not on the membership page and must be looked up per item from a different
source. This is independent of width and 2-D structure: a narrow \textsc{Easy} table can still be
\textsc{cross-source}, and overall the benchmark is 27 \textsc{exhaustive-only} to 201
\textsc{cross-source} (Figure~\ref{fig:composition}(b) breaks this split down by tier). My running
election grid is \textsc{cross-source}: the winners and parties come
off the results page, but each winner's age must be fetched from a separate biography. I find that
asking an LLM to predict this property from the question structure
alone is unreliable---roughly 72\% agreement with the web, including high-confidence errors---because
it mis-estimates what a given Korean page actually contains. I therefore web-ground the label:
an agent opens the best list page and verifies whether all requested columns are present, correcting
the structural guess.

\begin{table}[t]
\centering
\caption{Composition of \textbf{Ko-WideSearch} by difficulty tier.}
\label{tab:stats}
{\resizebox{\columnwidth}{!}{
\begin{tabular}{lrccrr}
\toprule
\textbf{Tier} & \textbf{$N$} & \textbf{Cols} & \textbf{2-D} & \textbf{Rows} & \textbf{Cells} \\
 & & \footnotesize med (range) & & \footnotesize med (range) & \\
\midrule
\textsc{Easy}   & 116 & 3 (2--4) & 0\%   & 14 (8--39) & 3{,}901 \\
\textsc{Medium} &  67 & 5 (3--8) & 30\%  & 16 (8--44) & 5{,}102 \\
\textsc{Hard}   &  45 & 7 (5--8) & 100\% & 21 (8--46) & 5{,}557 \\
\midrule
\textbf{Total}  & \textbf{228} & --- & --- & --- & \textbf{14{,}560} \\
\bottomrule
\end{tabular}}}{}
\end{table}

\subsection{Quality Control}
\paragraph{Question--gold consistency.} When the cross-check drops a source-fragile column, the
question is rewritten to ask only for the surviving columns, so the question never requests a cell that
has no gold value.
\paragraph{De-duplication and diversity.} I remove near-duplicate tables (same membership and
attributes) and cap per-entity repetition; the released benchmark spans 190 distinct set-parent
entities over its 228 tables, with per-entity counts reported to monitor over-representation.
\paragraph{Decontamination.} I screen every question against 5{,}744 questions from eight existing
evaluation sets---WideSearch, K-BrowseComp, BrowseComp-ZH, BrowseComp-Plus, GAIA, FRAMES, HLE, and
MCP-Atlas---using CJK-aware shingle Jaccard, $n$-gram containment, and answer overlap. No question
trips any of these three signals at the 0.6 threshold, confirming novelty.
\paragraph{Human spot-check.} A native-speaker reviewer audits a stratified sample across the three tiers
for membership completeness, boundary clarity, cell accuracy, and temporal stability; I position the
benchmark as pipeline-verified, with no external paid annotation and no formal inter-annotator
study.

\subsection{Dataset Statistics}
Table~\ref{tab:stats} gives the per-tier shape. Coverage spans sixteen categories and 190
distinct set-parent entities: 83\% of the 228 tasks enumerate a unique set, so even though sports/games leads, the
benchmark is diverse at the entity level. The 80 sports tasks are nearly all different leagues,
seasons, and tournaments, and the \textsc{Medium} tier alone covers fourteen categories, a topical
breadth that echoes that reported for K-BrowseComp~\cite{lee2026kbrowsecomp}. One skew is worth
flagging up front: the \textsc{Hard}, 2-D tier is sports-season heavy (30 of 45 tables). A second, orthogonal property
is each table's sourcing (\textsc{exhaustive-only} vs.\ \textsc{cross-source}). Overall the split is 27:201, and it
tracks difficulty: every single-page table is \textsc{Easy}, while \textsc{Medium} and \textsc{Hard}
are entirely cross-source, so a harder task always stitches its attributes across pages. The full
record schema and per-tier sample tasks are given in the appendix.

\section{Experimental Setup}
\paragraph{Agent harness.} I evaluate each model as a browsing agent over a single
\texttt{search}/\texttt{open}/\texttt{find} tool namespace pointed at Korean web search, under a fixed
per-question budget of thirty agent iterations and a single attempt per task (pass@1); each iteration
may batch several tool calls, so the total search-call count per task runs well above thirty. The agent
is told to enumerate the set exhaustively and to return exactly one structured table; my scorer then
parses that table directly (JSON, Markdown, or CSV, with a free-text fallback that yields recall only),
matches the predicted rows to the gold on the key column(s), and grades every cell with one shared,
deterministic type-aware comparator held constant across systems---so no model is advantaged or
penalized by an opaque judge.

\paragraph{Models.} I evaluate three families of systems, chosen so the comparison lines up with the
roster reported for K-BrowseComp~\cite{lee2026kbrowsecomp}: proprietary frontier models (e.g.\ the GPT, Claude,
and Gemini lines~\cite{openai2023gpt4,gemini2023}), open-weight models (e.g.\ GLM, Llama, Qwen, and
DeepSeek~\cite{grattafiori2024llama3,qwen2024qwen25,deepseek2024v3}), and Korean-specialized models
(e.g.\ EXAONE~\cite{lgai2024exaone}, HyperCLOVA~X~\cite{yoo2024hyperclovax}, A.X, and
Kanana~\cite{kakao2025kanana}). I score twenty systems end-to-end on the full pool---GPT-5.5,
Claude-Opus-4.8/4.7/4.6, Claude-Sonnet-4.6, Claude-Haiku-4.5, Gemini-3.1-Pro, Gemini-3.1-Flash-Lite,
Gemini-3.5-Flash, GPT-5.4, and GPT-5.4-mini/nano (proprietary); DeepSeek-V4-Pro, DeepSeek-Chat,
GLM-5.1, Gemma-4-31B, and Qwen3.6-35B (open-weight); and A.X-4.0, Solar-Open-2-preview, and K-EXAONE-236B (Korean-specialized).

\begin{figure*}[t]
\centering
\includegraphics[width=\textwidth]{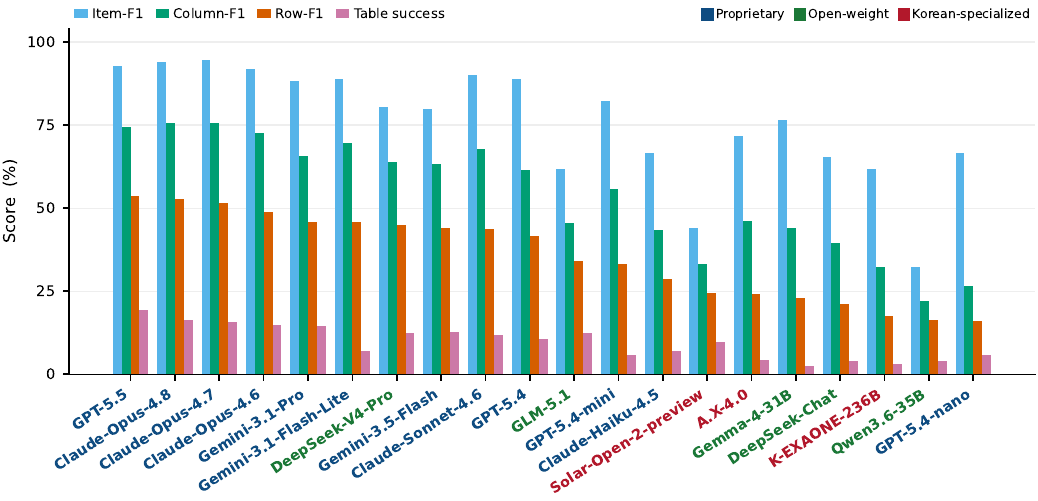}
\caption{\textbf{Membership is recovered; full rows and whole tables are not.} Item-F1 (membership),
Column-F1 (matched cells), Row-F1 (full rows), and table success (the whole table exactly correct), as
percentages, on all 228 tables for the twenty-system roster (sorted by Row-F1; x-axis labels colored by
family). Each model's bars cascade downward: the set (Item-F1, up to 94) is recovered far more
fully than whole rows (Row-F1 16--54), and even the strongest solves under a fifth of tables
outright. The bottleneck is exhaustive per-cell filling, not finding the membership; notably, the
open-weight \textbf{DeepSeek-V4-Pro} (Row-F1 45.0) stays competitive, mid-pack among the proprietary systems.}
\label{fig:results_gap}
\end{figure*}

\paragraph{Metrics.} I report the four WideSearch metrics---mean Item-, Column-, and Row-F1 and the
table-success rate---alongside the structured-output parse rate, and break each one down by difficulty
tier, by sourcing label, and by category. Item-F1 isolates membership: on the running election grid,
whether all thirty-four (province, round) rows are recovered, none missed and none invented. Row-F1 and
table success then add that every attribute cell---in a row, and across the whole table---must also be
correct. The parse rate captures whether a system can emit a readable table at all.

\section{Results}

\begin{table}[t]
\centering
\caption{Main results on \textbf{Ko-WideSearch} (228 tables), as percentages: the four WideSearch
metrics and the structured-output parse rate, by system family, single-pass (pass@1). Best per column in \textbf{bold}.}
\label{tab:main}
{\resizebox{\columnwidth}{!}{
\begin{tabular}{lccccc}
\toprule
\textbf{Model} & \textbf{Item-F1} & \textbf{Column-F1} & \textbf{Row-F1} & \textbf{Tab.\,Succ.} & \textbf{Parse} \\
\midrule
\multicolumn{6}{l}{\textit{Proprietary frontier}} \\
GPT-5.5 & 92.8 & 74.3 & \textbf{53.7} & \textbf{19.3} & 98.2 \\
Claude-Opus-4.8 & 94.1 & 75.5 & 52.9 & 16.2 & 99.6 \\
Claude-Opus-4.7 & \textbf{94.6} & \textbf{75.6} & 51.6 & 15.8 & \textbf{100.0} \\
Claude-Opus-4.6 & 92.0 & 72.7 & 48.9 & 14.9 & 100.0 \\
Gemini-3.1-Pro & 88.2 & 65.6 & 45.9 & 14.5 & 93.0 \\
Gemini-3.1-Flash-Lite & 89.0 & 69.7 & 45.9 & 7.0 & 99.1 \\
Gemini-3.5-Flash & 79.9 & 63.3 & 44.1 & 12.7 & 86.8 \\
Claude-Sonnet-4.6 & 90.2 & 67.7 & 43.6 & 11.8 & 100.0 \\
GPT-5.4 & 89.0 & 61.5 & 41.6 & 10.5 & 100.0 \\
GPT-5.4-mini & 82.3 & 55.9 & 33.3 & 5.7 & 98.2 \\
Claude-Haiku-4.5 & 66.7 & 43.5 & 28.8 & 7.0 & 75.9 \\
GPT-5.4-nano & 66.6 & 26.6 & 15.9 & 5.7 & 93.4 \\
\midrule
\multicolumn{6}{l}{\textit{Open-weight}} \\
DeepSeek-V4-Pro & 80.4 & 63.9 & 45.0 & 12.3 & 87.3 \\
GLM-5.1 & 61.7 & 45.6 & 34.0 & 12.3 & 66.2 \\
Gemma-4-31B & 76.4 & 43.9 & 23.0 & 2.6 & 93.4 \\
DeepSeek-Chat & 65.4 & 39.4 & 21.3 & 4.0 & 87.3 \\
Qwen3.6-35B & 32.4 & 22.2 & 16.2 & 4.0 & 38.6 \\
\midrule
\multicolumn{6}{l}{\textit{Korean-specialized}} \\
Solar-Open-2-preview & 44.0 & 33.3 & 24.4 & 9.7 & 62.7 \\
A.X-4.0 & 71.7 & 46.2 & 24.2 & 4.4 & 93.4 \\
K-EXAONE-236B & 61.9 & 32.3 & 17.5 & 3.1 & 82.9 \\
\bottomrule
\end{tabular}}}{}
\end{table}

\paragraph{RQ1: Is membership recovered while full rows are not?} Yes, and the gap is wide. Every system
in Table~\ref{tab:main} recovers most of the closed set yet completes far fewer full rows---the strongest
model, GPT-5.5, scores Item-F1 92.8 but Row-F1 53.7, and table success only 19.3 (about one table in
five entirely correct). Figure~\ref{fig:results_gap} traces this cascade per model: an agent that recovers the full
thirty-four-row election grid yet mis-fills a single turnout or a winner's age scores high on Item-F1 and
far lower on Row-F1. This is the WideSearch failure mode, now confirmed on
Korean sources. I want to highlight two things. First, the frontier is effectively tied: GPT-5.5,
Claude-Opus-4.8, and Claude-Opus-4.7 sit within two points on Row-F1 (53.7/52.9/51.6), and
Claude-Opus-4.7 in fact leads on membership and per-cell accuracy (Item-F1 94.6, Column-F1 75.6).
Second, the open-weight DeepSeek-V4-Pro (Row-F1 45.0) stays competitive with the frontier: it ranks
mid-pack among the proprietary systems---behind the Claude-Opus line and both Gemini-3.1 models, yet ahead of
GPT-5.4, Claude-Sonnet-4.6, and every smaller proprietary tier, so the best open model still outscores
half the proprietary field.

\begin{figure*}[t]
\centering
\includegraphics[width=\textwidth]{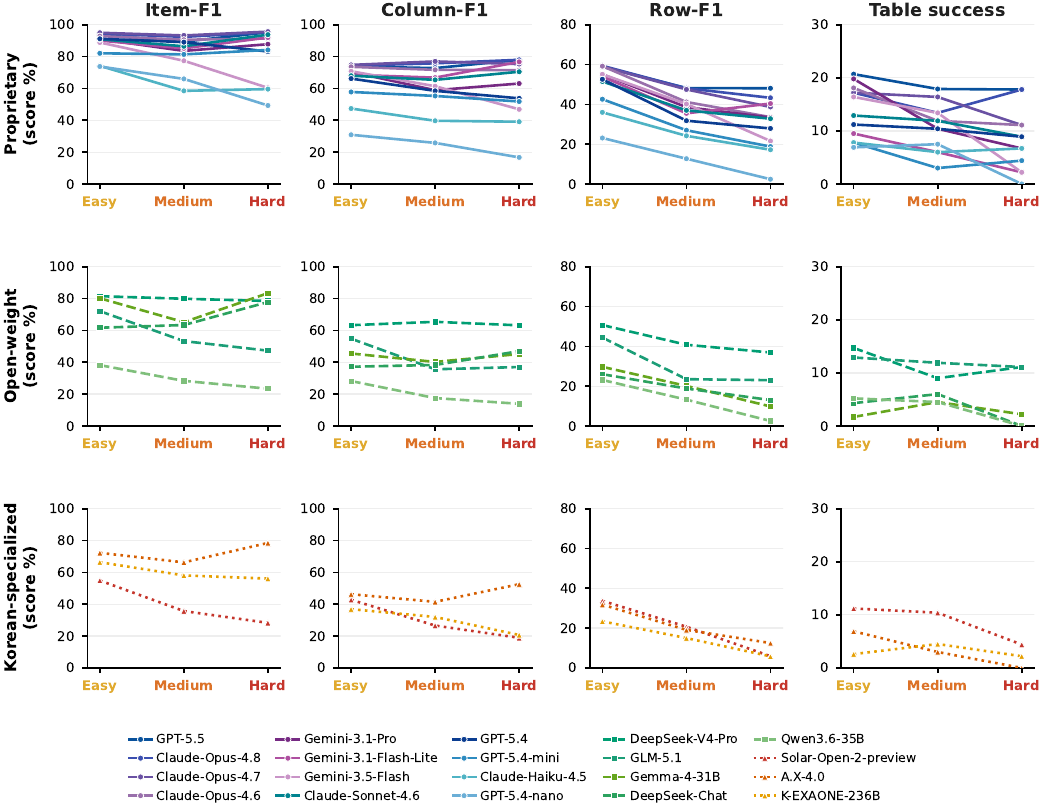}
\caption{\textbf{The metric breakdown by difficulty, faceted by model family.} Each of the four WideSearch
metrics (columns: Item-F1, Column-F1, Row-F1, and table success), as percentages, across the difficulty
tiers (\textsc{Easy}/\textsc{Medium}/\textsc{Hard}), with one row of panels per family---proprietary
(solid, top), open-weight (dashed, middle), Korean-specialized (dotted, bottom); $y$-scales are shared
down each column so the families are directly comparable. Faceting keeps all twenty models legible
rather than overplotting them in one panel. Membership (Item-F1) holds up across the tiers, but every
downstream metric falls as width and the 2-D composite key are added---Row-F1 and especially table
success drop steeply \textsc{Easy}$\rightarrow$\textsc{Hard}---and the family rows separate top to
bottom. The orthogonal sourcing breakdown is in Figure~\ref{fig:results_tiers_src} (Appendix); since the
27 exhaustive-only tables are all \textsc{Easy}, it partly re-expresses difficulty.}
\label{fig:results_tiers}
\end{figure*}

\begin{figure*}[t]
\centering
\includegraphics[width=\textwidth]{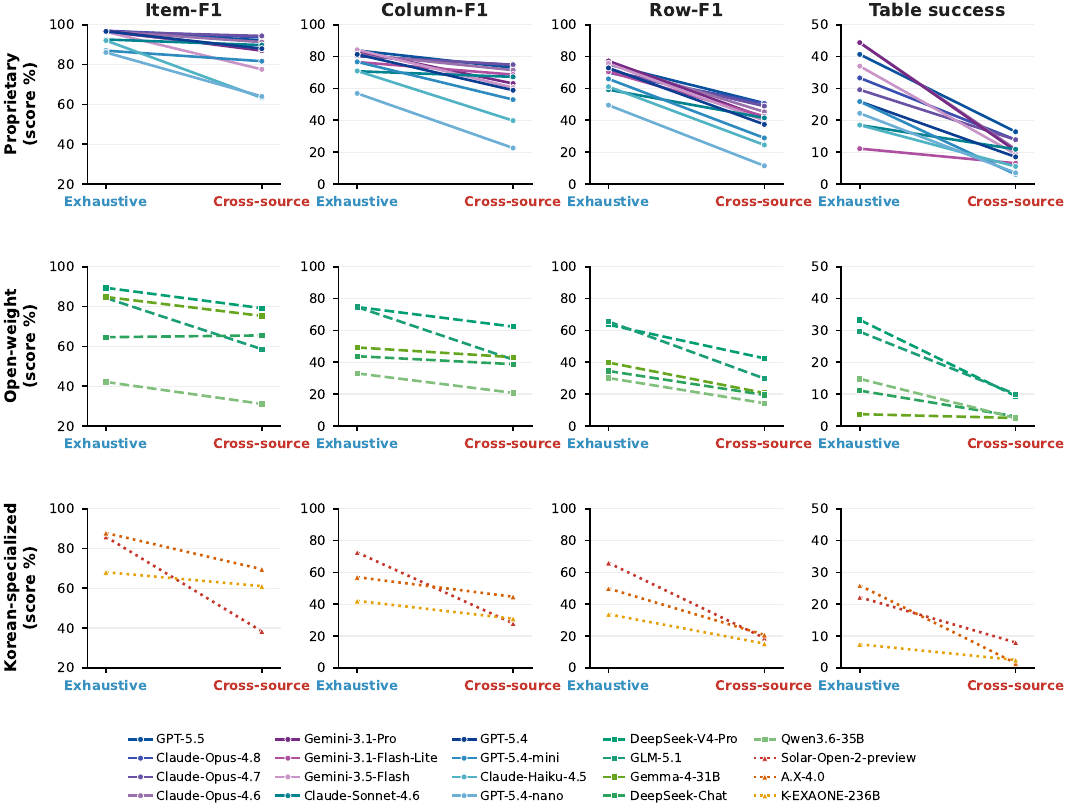}
\caption{\textbf{The metric breakdown by sourcing label, faceted by model family} (companion to
Figure~\ref{fig:results_tiers}). The same facet-by-family layout, but the $x$-axis is the orthogonal
sourcing label---single-page \textsc{exhaustive-only} versus \textsc{cross-source} tables. Every metric
drops from exhaustive to cross-source for nearly every model; because all 27 exhaustive-only tables are
\textsc{Easy}, this gap partly re-expresses difficulty rather than a fully independent axis.}
\label{fig:results_tiers_src}
\end{figure*}

\paragraph{RQ2: How steep is the difficulty gradient?} Steep, and along both structural axes
(Figure~\ref{fig:results_tiers} for difficulty; Appendix Figure~\ref{fig:results_tiers_src} for
sourcing). Along the difficulty axis, Row-F1 falls from
\textsc{Easy} to \textsc{Hard} for every model as width and the 2-D composite key are added---GPT-5.4-mini
from 42.6 to 18.8, DeepSeek-V4-Pro from 50.7 to 36.9---while the strongest model flattens near the top
(GPT-5.5, 58.9 to 48.1). Membership holds up across tiers: Item-F1 is roughly flat, and even rises on
\textsc{Hard}, where the sports-season grids make the set systematically enumerable---so the tier drop is
a row-completion effect, not a retrieval one. The orthogonal sourcing label shows the same
shape---single-page \textsc{exhaustive-only} tables are far easier than \textsc{cross-source} ones
(GPT-5.5 75.1 vs.\ 50.8), though, since the exhaustive-only tables are all \textsc{Easy}, that gap
partly reflects difficulty rather than sourcing alone.

\paragraph{RQ3: Do Korean-specialized models close the gap to frontier models?} No---the three I evaluate
sit at the open-weight floor. A.X-4.0 (Row-F1 24.2) and Solar-Open-2-preview (24.4) land level with
Gemma-4-31B (23.0), far below the frontier (GPT-5.5, 53.7) and the best open-weight model
(DeepSeek-V4-Pro, 45.0): native Korean fluency does not, on its own, overcome the benchmark's core demand
of long-horizon search followed by exhaustive per-cell filling. The two fail differently, and the
contrast is instructive. A.X-4.0 recovers membership about as well as a mid-tier open model (Item-F1
71.7) but cannot fill the cells (Row-F1 24.2); Solar-Open-2-preview's far lower Item-F1 (44.0), by contrast, is a
structured-output failure rather than a search one---it returns a scorable table only 62.7\% of the time,
often searching extensively and then emitting its findings as prose or a numbered list instead of the
requested table. K-EXAONE-236B lands even lower (Row-F1 17.5), at the open-weight floor, so all three
Korean-specialized systems sit far below the frontier (53.7) and the best open model (45.0); the
remaining Korean systems---HyperCLOVA~X and Kanana---are pending, so I read this as an early, partial
answer, to be completed against the Korean roster of K-BrowseComp~\cite{lee2026kbrowsecomp}.

\section{Analysis}

\begin{figure*}[t]
\centering
\includegraphics[width=\textwidth]{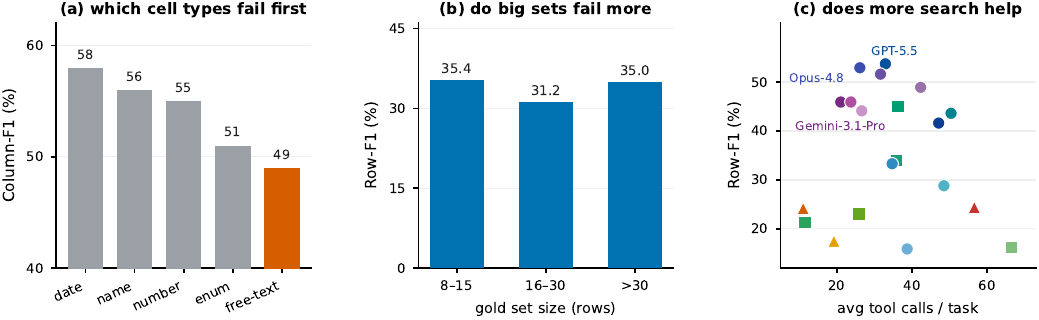}
\caption{\textbf{Where and why agents break.} (a) Per-cell-type Column-F1 on an instrumented
three-system subset (pooled): the open-ended \textbf{free-text} cells are filled least reliably and the
format-constrained ones (dates, names) most, so the difficulty is finding and normalizing the value, not
formatting it. (b) Row-F1 by gold set size on the full pool is essentially flat---breadth alone does not
drive failure. (c) Average tool calls per table against Row-F1: \textbf{more search does not buy
completeness}---the two heaviest searchers (Qwen3.6-35B, Solar-Open-2-preview) score lowest, while GPT-5.5 and
Claude-Opus-4.8 lead the benchmark with moderate search. Marker shapes: circle (proprietary), square
(open-weight), triangle (Korean-specialized); model colors as in Figure~\ref{fig:results_tiers}.}
\label{fig:analysis}
\end{figure*}

\paragraph{Which attribute types fail first?} Free-text attributes, not the structured ones. Breaking
Column-F1 down by declared cell type on an instrumented subset---three reference systems (GPT-5.5,
DeepSeek-V4-Pro, GPT-5.4-mini) re-run with per-column scoring---pooled accuracy falls from declared
dates (58) and names (56) through numbers (55) and enums (51) to
free-text cells (49): the open-ended attributes an agent must read off a page and paraphrase are
filled least reliably, while the format-constrained ones are easiest (Figure~\ref{fig:analysis}(a)). The ordering locates the
difficulty in finding and normalizing the value, not formatting it---my comparator already tolerates
surface form (date granularity, thousands-commas, name variants), so a low free-text score is a wrong or
unverified value rather than a mis-format.

\begin{figure}[t]
\centering
\includegraphics[width=\columnwidth]{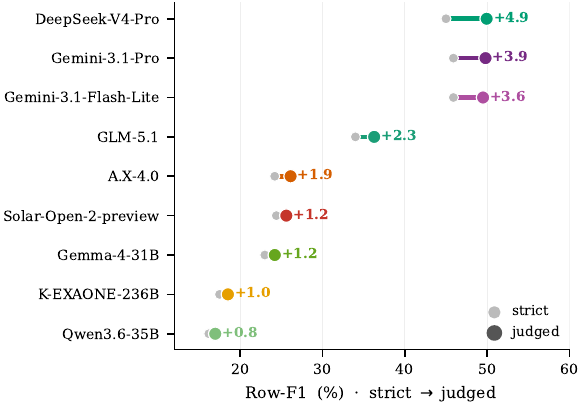}
\caption{\textbf{Semantic judging helps strong models more.} Strict $\rightarrow$ judged Row-F1 per
model (full 228-table pool). The dumbbell length is the correction the LLM judge applies; it grows with
model strength (DeepSeek-V4-Pro $+4.9$ at top vs.\ $+0.8$ for the weakest), so strict cell-matching
under-states the gap between strong and weak systems.}
\label{fig:judged_delta}
\end{figure}

\begin{figure*}[t]
\centering
\includegraphics[width=\textwidth]{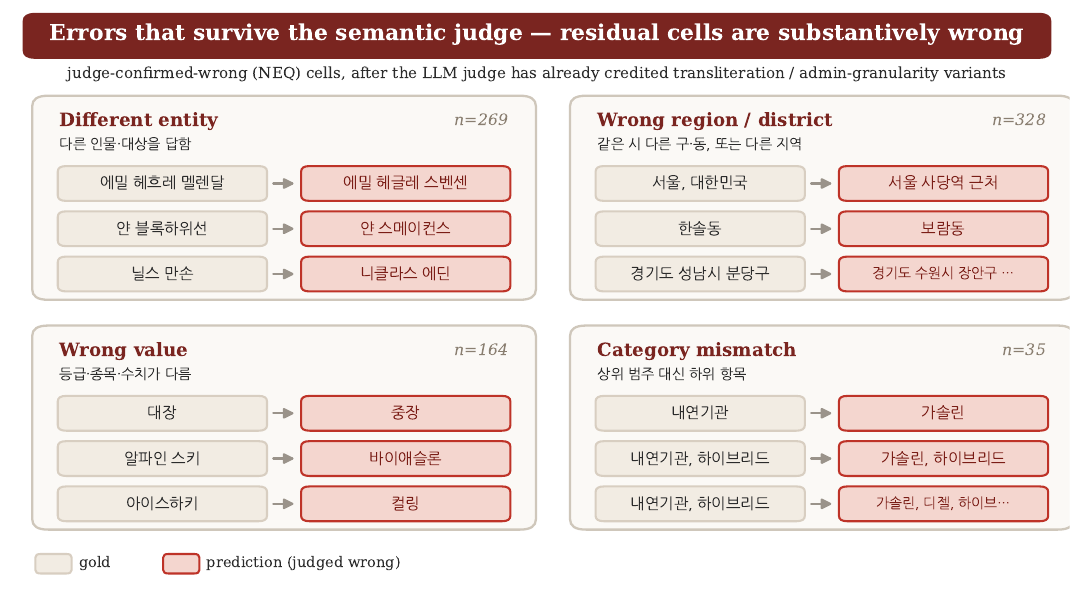}
\caption{\textbf{Errors that survive the semantic judge.} After the judge credits transliteration and
administrative-granularity variants, the residual judge-confirmed-wrong cells are substantive: a
different entity ($n{=}269$), a different region/district ($328$), a wrong value ($164$), or a category
mismatch ($35$). The Item-F1$\,\gg\,$Row-F1 gap is factual error, not formatting.}
\label{fig:neq_residual}
\end{figure*}

\paragraph{Does strict cell-matching under-state the gap?} That last claim rests on the comparator
tolerating surface form; to measure how much it still misses, I add a second, semantic pass. For every
cell the deterministic comparator scores wrong on a soft-typed column (name, location, free text), an
LLM judge (GPT-5.4-mini) decides whether the prediction nonetheless denotes the same answer as the
gold---crediting transliteration variants and administrative granularity (``Gangwon Chuncheon''
$\equiv$ ``Chuncheon, Korea''), but not a different entity, a different value, or fabricated
specificity. Re-scoring this way (Table~\ref{tab:judged},
Figure~\ref{fig:judged_delta}) moves Row-F1 up by only $0.8$--$4.9$ points, and the gain rises with
model strength: the strongest systems gain most (DeepSeek-V4-Pro $+4.9$, Gemini-3.1-Pro $+3.9$,
Gemini-3.1-Flash-Lite $+3.6$), while the weakest (Qwen3.6, K-EXAONE) gain $0.8$--$1.0$. The reason is in the residual: among a model's
judge-confirmed-wrong cells, the share the judge rescues rises with model strength (GPT-5.5 $35\%$,
K-EXAONE $9\%$), so a strong model's remaining errors are disproportionately surface variants while a
weak model's are substantive---a different district, rank, or person (Figure~\ref{fig:neq_residual}).
Strict matching therefore under-states the gap rather than inflating it: semantic judging
separates strong from weak slightly more, not less. I keep the deterministic scorer as the reproducible
default (Table~\ref{tab:main}) and report these judged numbers only as a measurement-validity check---the
judge is a single cheap model grading, among others, its own family, so I read the absolute judged values
as a lower-bound correction, not a new ranking.

\begin{table}[t]
\centering
\caption{Semantic-judged vs.\ strict Row-F1 (\%): an LLM judge credits
same-referent surface variants (transliteration, admin-granularity) on soft-typed cells that the
deterministic scorer marks wrong. \textbf{Row str.}\ is the strict Row-F1 from Table~\ref{tab:main};
$\Delta$ is the judge correction measured on each model's logged predictions; \textbf{Row jdg.}~$=$~str.~$+\,\Delta$.
}
\label{tab:judged}
{\small
\begin{tabular}{lccc}
\toprule
\textbf{Model} & \textbf{Row str.} & \textbf{Row jdg.} & \textbf{$\Delta$} \\
\midrule
Gemini-3.1-Pro & 45.9 & 49.8 & $+3.9$ \\
Gemini-3.1-Flash-Lite & 45.9 & 49.5 & $+3.6$ \\
DeepSeek-V4-Pro & 45.0 & \textbf{49.9} & $+4.9$ \\
GLM-5.1 & 34.0 & 36.3 & $+2.3$ \\
Solar-Open-2-preview & 24.4 & 25.6 & $+1.2$ \\
A.X-4.0 & 24.2 & 26.1 & $+1.9$ \\
Gemma-4-31B & 23.0 & 24.2 & $+1.2$ \\
K-EXAONE-236B & 17.5 & 18.5 & $+1.0$ \\
Qwen3.6-35B & 16.2 & 17.0 & $+0.8$ \\
\bottomrule
\end{tabular}}
\vspace{-6pt}
\end{table}

\begin{figure}[t]
\centering
\includegraphics[width=\columnwidth]{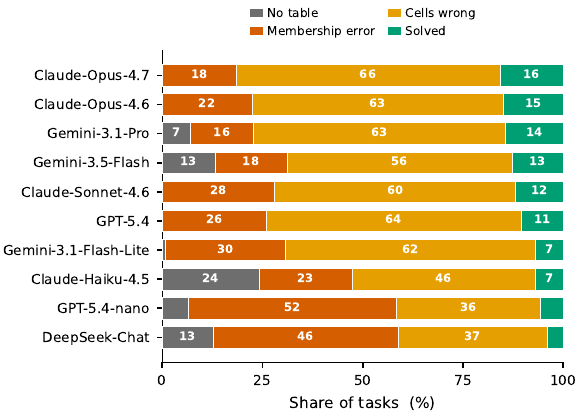}
\caption{\textbf{Failure composition by stage}, for the ten timely-routed systems with per-task logging.
Every task is assigned to its first failure stage---no parseable table, a substantially wrong row set
(\emph{membership}, item-F1 $<0.9$), an essentially-correct set with a wrong attribute cell
(\emph{cells}), or fully \emph{solved}---so the four shares sum over the 228 tasks. For capable systems
the bottleneck is cell-filling (cells 56--66\%); the smallest fail a stage earlier at membership
(46--52\%); parse failure is rare except for \textbf{Claude-Haiku-4.5} (24\%).}
\label{fig:error_types}
\end{figure}


\begin{figure*}[t]
\centering
\includegraphics[width=\textwidth]{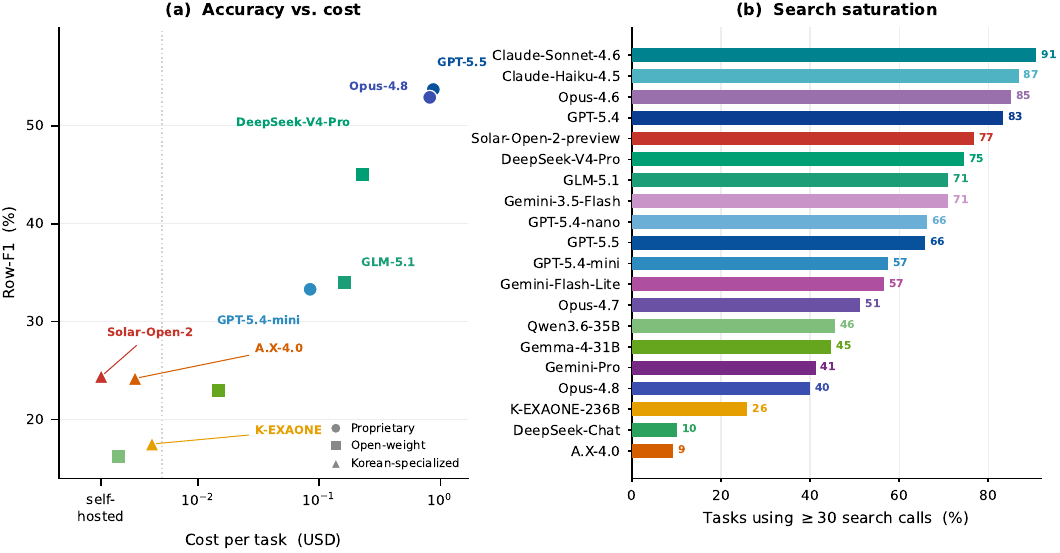}
\caption{\textbf{Neither dollars nor search calls close the gap.} \textbf{(a)} Row-F1 against cost per
task for the six API-served models with metered cost. Accuracy rises with cost then flattens: the
open-weight \textbf{DeepSeek-V4-Pro} reaches most of the frontier's Row-F1 at a fraction of the price of
GPT-5.5 and Claude-Opus-4.8. \textbf{(b)} Share of
tasks on which a model spends at least thirty tool calls, over all twenty systems; the models that cross
that bar most often (Claude-Sonnet-4.6, Claude-Haiku-4.5) are not the top scorers, while the top systems
search far less. The per-question budget is thirty agent iterations,
each able to batch several calls, so totals run higher---to 947 for Qwen3.6. The four self-hosted
models (A.X-4.0, Solar-Open-2-preview, Qwen3.6-35B, K-EXAONE-236B) have no API fee, so they sit at the leftmost
``self-hosted'' position of (a) at their Row-F1 (their small horizontal spread is jitter, not cost).}
\label{fig:cost_budget}
\end{figure*}

\paragraph{Failure taxonomy.} The recurring failure is cell-filling, not set-boundary confusion.
Decomposing membership into precision and recall on the same subset, the two are balanced for every
system---GPT-5.5 recovers the set at precision 85 and recall 86, DeepSeek-V4-Pro at 71 and 71---so agents
neither systematically invent extra members nor systematically drop them; the closed set is found
cleanly. The collapse is downstream: whole-row precision and recall both fall to 25--37, the signature of
a recovered row carrying one wrong cell. A second, distinct mode is parse failure---a system
searches, sometimes even finds the answer, but returns it as prose or a numbered list instead of the
requested table and scores zero outright. This is rare at the frontier (GPT-5.x parse rate 98) but common
elsewhere (DeepSeek-V4-Pro 87, and my own Solar-Open-2-preview at 63---its dominant failure). As in WideSearch,
the errors are an exhaustiveness problem, and they concentrate in the harder tiers
(Figure~\ref{fig:results_tiers}).

\paragraph{Failure composition shifts with capability.} Assigning every task to its first failure
stage---no parseable table, a substantially wrong set, an essentially-correct set with a wrong cell, or
solved---confirms the cascade per model (Figure~\ref{fig:error_types}). For the capable systems the
bottleneck is squarely cell-filling: \textbf{Claude-Opus-4.7} recovers the set on all but 18\% of tasks,
yet a single wrong cell costs it 66\% and only 16\% are fully solved. The smallest systems fail a stage
earlier---\textbf{GPT-5.4-nano} and \textbf{DeepSeek-Chat} get the row set itself wrong on 46--52\% of
tasks, missing or inventing rows before cells are at issue. Parse failure stays rare except for
\textbf{Claude-Haiku-4.5}, which returns no scorable table on 24\%. Capability resolves set-finding, but
the per-cell ceiling holds: no system clears more than 16\% of whole tables.

\paragraph{Set size and search effort.} Two intuitions about breadth turn out to be wrong here. First,
larger sets are not harder: pooling Row-F1 by gold set size, accuracy is essentially flat from small to
large tables (35.4 at 8--15 rows, 31.2 at 16--30, 35.0 beyond 30; Figure~\ref{fig:analysis}(b))---the biggest sets are often
systematically enumerable sports seasons, so set size alone does not drive failure; width and the
2-D key do. Second, and more striking, more search does not buy completeness. I want to highlight that
the two systems that search hardest---Qwen3.6 at 66 tool calls per table and Solar-Open-2-preview at 57---are the
lowest- and near-lowest-scoring (Row-F1 16 and 24), while GPT-5.5 and Claude-Opus-4.8 reach the top with
moderate search (33 and 26 calls; Figure~\ref{fig:analysis}(c)). The ceiling is therefore
the agent's ability to assemble and verify a table, not its search budget; the heavy searchers thrash
rather than converge.

\paragraph{Neither dollars nor more search close the gap.} The same ceiling appears when I price the
search. Figure~\ref{fig:cost_budget}(a) plots Row-F1 against cost per task, and accuracy saturates
steeply: GPT-5.5 spends about \$0.87 a table for Row-F1 53.7, but DeepSeek-V4-Pro reaches 45.0 at a
quarter of that (\$0.23), so the last nine points over the best open model cost roughly an order of
magnitude more.
Effort tells the same story (Figure~\ref{fig:cost_budget}(b)): Claude-Sonnet-4.6 crosses thirty tool
calls on 91\% of tasks for a mid-pack Row-F1 43.6, while Claude-Opus-4.8 reaches second place (52.9)
doing so on only 40\%. More money and more tool calls do not buy filled cells.

\paragraph{The leaderboard is stable.} These gaps are not sampling noise. Re-running three systems three
times each on a thirty-task subset (pass@1, temperature 0.7), Row-F1 standard deviation is 0.016
(DeepSeek-V4-Pro), 0.025 (GPT-5.4-mini), and 0.040 (GLM-5.1), so differences above about five points are
robust. This is also why I read RQ3 as a tie rather than a ranking: the A.X-4.0, Solar-Open-2-preview, and
Gemma-4-31B cluster (23.0--24.4 Row-F1) sits well inside that noise band.

\paragraph{Table success is the truest metric but the least discriminating.} Table success---the whole
table exactly right, WideSearch's pass criterion---is the outcome that matters, and the starkest: even
GPT-5.5 clears under a fifth of tables. But it saturates near the floor, where most systems sit in single
digits, so it separates the field far less than Row-F1's 15.9--53.7 spread.
Being all-or-nothing per table, it also rewards getting small, easy tables perfectly: Solar-Open-2-preview's
table success (9.7) tops GPT-5.4-mini's (5.7), even though its
Row-F1 (24.4) trails it (33.3)---an inversion from nailing a few short tables while collapsing
on the rest. I therefore headline table success as the end-to-end outcome but rank systems by Row-F1,
which credits every correct cell and is robust to table size.

\section{Conclusion and Limitations}
I introduced \textsc{Ko-WideSearch}, a Korean breadth-search benchmark of 228 tables in three
difficulty tiers, built by an automated synthesize-and-verify pipeline, scored by a
normalization-aware comparator shared between gold construction and grading, and labeled with
web-grounded sourcing tiers. On a twenty-model roster, the headline is clear: agents
recover the membership but not the rows (GPT-5.5 reaches Item-F1 92.8 against Row-F1 53.7), and accuracy
falls steadily as my width and 2-D-key knobs harden the tier. Breadth search is hardest to evaluate where it matters most: a
complete, per-cell-correct gold set is expensive to build by hand, and outside English it barely
exists. However, breadth is also what makes a gold table checkable---its set can be
independently re-enumerated and its cells re-looked-up---so construction can be automated and verified
rather than hand-curated. I therefore believe an automated, verifiable pipeline is the practical way
to keep breadth-search evaluation current as the web shifts, and I open-source the pipeline and scorer
so the benchmark can be re-grown. One caveat shapes distribution: because the tasks are solved on the live
web, I release the evaluation data by request rather than posting it openly, keeping the benchmark a
genuine search task.

\paragraph{Limitations.} My coverage has several edges. One is category skew: the \textsc{Hard}, 2-D
tier is still sports-season heavy (roughly 67\%), because a compact season table is the most reliably
buildable cross-product and non-sports 2-D coverage (e.g.\ elections) is only partial. Also, each table's
source is anchored at the primary membership page---one source URL per table rather than per attribute---though
cross-source verification was applied at build time, and performance is measured under a single
harness, search backend, and budget. Finally, because web evidence and rankings shift over time,
volatile tables carry an ``as-of'' date and need periodic re-validation, and the sourcing-tier label
reflects page structure at labeling time. However, my pipeline is built precisely so that this
re-validation and regrowth are automated rather than manual, and I position the benchmark as
pipeline-verified with a native-speaker spot-check rather than as externally gold-annotated.

\section{Technical Appendix}

\subsection{Data Schema}
Each released task is one JSON record. It carries an \texttt{id} and a \texttt{difficulty\_tier}
(\textsc{easy}/\textsc{medium}/\textsc{hard}); a \texttt{category} and the orthogonal
\texttt{hardness\_tier} (the \textsc{exhaustive\_only}/\textsc{cross\_source} sourcing label); an
\texttt{as\_of} date for volatile tables (null otherwise); the natural-language \texttt{question}; the
ordered \texttt{columns} with per-column \texttt{column\_specs} (a format such as
\texttt{name}/\texttt{int}/\texttt{float:N}/\texttt{date:YYYY-MM-DD}/\texttt{enum:A|B}); the
\texttt{key\_columns} that jointly identify a row (one for a primary key, two for a 2-D grid); the gold
\texttt{answer\_set} of \{\texttt{name}, \texttt{attrs}\} rows, where a sentinel marks a
genuinely-absent cell; any \texttt{exclusions}; \texttt{n\_rows} and \texttt{n\_cols}; the
authoritative \texttt{sources}; and an \texttt{evaluation} contract giving \texttt{unique\_columns},
the \texttt{required} columns, and a per-column \texttt{eval\_pipeline} (each column's preprocessing
and metric).

\subsection{Construction and Verification Details}
The gold tables are built and verified by frontier models (GPT-5.4 and DeepSeek-V4-flash) issuing live
Korean web searches over the \texttt{search}/\texttt{open}/\texttt{find} namespace. The three
acceptance gates are applied independently: non-memorizability fails closed, so an inconclusive
closed-book check counts as a rejection and a table whose closed-book cell recall reaches $0.5$ is
rejected; completeness requires an independent re-enumeration of the membership to match the gold at
set-F1 $\geq 0.7$, run on a different model family than the builder so set agreement is not
self-confirming; and cross-source verification re-looks-up each attribute and drops any column that
agrees with the independent source below $0.6$, requiring at least one attribute column to survive. The
two difficulty knobs are set as generation targets: width (tables with $\geq 5$ columns) and a 2-D
composite key (\texttt{key\_columns} of length two, so membership is a cross-product); turning both
yields the \textsc{Hard} tier. A closed set too large to enumerate in one pass
(100+ members) is built by partitioning it into bounded sub-ranges, enumerating each independently and
taking the union, with the partition keys themselves re-enumerated and set-matched so a missed
sub-range is caught.

\subsection{Pipeline Prompts}
All construction, verification, and evaluation prompts ship with the code; the build and verification
prompts are issued in English, the evaluation and labeling prompts in Korean. I render each faithfully
below (Korean prompts in English).

\paragraph{Build agent.} The builder is told to design a hard ``wide'' search task whose answer is
the complete set of entities satisfying a precise predicate---breadth, not a single answer---and to
assemble the gold set by exhaustive \texttt{search}/\texttt{open}/\texttt{find}. Its requirements: the
predicate must define a closed, finite, officially-bounded set with an authoritative complete list;
target 8--40 members (restrict the predicate to a sub-range if larger); the primary key is each item's
canonical name, never a positional placeholder; the hardness must come from a wide, multi-source
table of 4--7 attribute columns whose values live on different pages; the columns must be heterogeneous
(a date, a number, a name, and where natural an enum, money, location, or URL), each carrying an explicit
per-cell format directive embedded in the question; volatile columns are pinned to an as-of date; and the
answer must not be obtainable from a single page or from memory. For Korean, two nudges are appended:
prefer the larger complete set within the window (toward 25--40 members), and when the enumeration is
naturally a cross-product of two dimensions (team\,$\times$\,season), emit both dimensions as the first
two columns and list them in \texttt{key\_columns}. The builder returns a JSON object with the question,
predicate, columns, \texttt{column\_specs}, \texttt{key\_columns}, exclusions, the enumerated items, and
evidence URLs.

\paragraph{Verification gates.} The three prompts mirror the acceptance logic above.
\textbf{(1) Non-memorizability:} a closed-book model (told it has no web access) reproduces the answer
table from memory, and a judge scores the fraction of gold cells---item and attribute, not just
names---it recovers. \textbf{(2) Completeness:} an independent agent re-enumerates the membership from
the question alone, and a set-match judge reports agreement in both directions. \textbf{(3) Cross-source
attribute verification:} a fact-checker independently re-looks-up each attribute---told not to trust any
value it is given---and reports each cell in the column's canonical format.

\begin{table}[t]
\centering
\caption{Pooled metrics (mean across the twenty systems, \%) by difficulty tier and by sourcing label.
Row-F1 and table success fall steeply with difficulty and with cross-sourcing, while membership
(Item-F1) holds up; since all 27 \textsc{exhaustive-only} tables are \textsc{Easy}, the sourcing split
partly re-expresses difficulty rather than a fully independent axis.}
\label{tab:pooled}
{\resizebox{\columnwidth}{!}{
\begin{tabular}{lcccc}
\toprule
 & \textbf{Item-F1} & \textbf{Column-F1} & \textbf{Row-F1} & \textbf{Tab.\,Succ.} \\
\midrule
\textsc{Easy}   & 79.3 & 56.6 & 43.5 & 11.3 \\
\textsc{Medium} & 72.2 & 51.3 & 30.2 & 9.1 \\
\textsc{Hard}   & 73.1 & 51.0 & 23.4 & 6.4 \\
\midrule
\textsc{Exhaustive-only} & 86.9 & 68.7 & 60.4 & 25.0 \\
\textsc{Cross-source}    & 74.5 & 51.9 & 32.3 & 7.6 \\
\bottomrule
\end{tabular}}}{}
\vspace{-6pt}
\end{table}

\begin{figure}[t]
\centering
\includegraphics[width=\columnwidth]{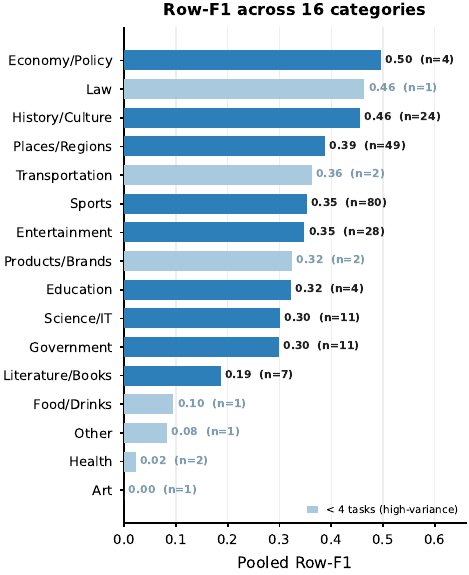}
\caption{Pooled Row-F1 by topical category (mean across the twenty systems, model-task weighted). Most
categories sit near the 0.36 pooled mean; Literature/Books is the clear low outlier and the
largest category, Sports, sits mid-pack. Small-sample categories ($<4$ tasks) are faded as high-variance.}
\label{fig:percategory}
\end{figure}

\paragraph{Evaluation harness.} Each evaluated model receives a Korean system prompt to act as a
web-search agent over \texttt{search}/\texttt{open}/\texttt{find}, to find the complete set without
omitting any item and fill each item's requested attributes, to write only search-confirmed values (no
guessing), and to leave unknown cells blank; the user turn appends the question and requires the answer
to end with exactly one JSON block whose \texttt{attrs} keys are the requested attribute columns.

\paragraph{Sourcing-tier labeling.} A structural classifier first labels each table
\textsc{exhaustive-only} or \textsc{cross-source} from the question alone---whether the complete answer
usually sits on one Korean page or needs separate per-item pages---defaulting to \textsc{cross-source}
with lower confidence when unsure. A web-grounded calibrator then opens the best list page and reports
whether that one page carries every requested column, overwriting the structural guess; the comparison
yields the roughly 72\% structural-accuracy figure.

\paragraph{Verbalizer and decontamination.} A verbalizer rewrites the bare imperative question into a
natural persona-and-motivation scenario that still demands every item and fixes the output format,
without changing the set, columns, as-of date, or scope. Decontamination uses no model: it screens every
question against the eight existing benchmarks with exact-normalized match, MinHash/Jaccard shingle
overlap, and contiguous $n$-gram containment (CJK character 4-grams, ASCII word-grams) at a $0.6$
threshold.

\subsection{Scoring Details}
The scorer parses the model's final answer into rows---a JSON \texttt{items} block, a Markdown/pipe
table, or CSV; if none parse, a free-text fallback yields recall only and the table scores zero (a
parse failure). Predicted rows are matched one-to-one to the gold on the key column(s)---a single
primary key, or the pair of dimensions for a 2-D grid---so precision, recall, and F1 never exceed one
even when keys legitimately repeat. Each non-sentinel gold cell is compared with one shared type-aware
comparator: dates at their common granularity (a year matches a full date with the same year), numbers
after stripping thousands separators and units with a 5\% relative tolerance (the AGGBench count
tolerance), URLs by host and path, and names or locations by normalized text with substring and token
overlap. The same comparator builds the gold and grades the model, so a stable column is neither dropped
at construction nor mis-scored at evaluation. The three reported metrics follow: Item-F1 is membership over keys; Column-F1 is per-attribute
cell agreement over matched rows; Row-F1 requires the key and every attribute cell; table success is
Row-F1 $=1$ across the whole table.

\subsection{Sample Tasks}
I give representative tasks across the three tiers; the questions are translated from Korean, and the
originals with full gold tables ship in the released \texttt{SAMPLES.md}. Each task lists its per-column
comparison types; free-text columns are graded by the deterministic normalized-text match, not a judge.
\begin{itemize}
\item \textbf{\textsc{Easy}} (\texttt{kws-0008}). ``List all eight planets of the Solar System
with each planet's discovery year and discoverer.'' Key \texttt{[planet]}; cells scored as: planet
(exact match), discovery year and discoverer (free-text). 8 rows, 3 columns.
\item \textbf{\textsc{Medium}} (\texttt{kws-mh001}). ``List every low-cost carrier operating in
Korea as of 2026-06, with parent company, launch year, main hub, and current fleet size.'' Key
\texttt{[carrier]}; cells: carrier (exact), parent company and hub (free-text), launch year (date
tolerance), fleet size (numeric tolerance). 9 rows, 5 columns.
\item \textbf{\textsc{Hard}} (\texttt{kws-mh004}). ``For all 17 metropolitan provinces and
cities, list the head (mayor or governor) elected in the 7th (2018) and 8th (2022) local elections,
with party, voter turnout, the winner's age, and vote share.'' Key \texttt{[province, election round]}
(a 2-D grid); cells: keys (exact match), winner and party (free-text), turnout, age, and share
(numeric tolerance). 34 rows, 7 columns.
\item \textbf{\textsc{Easy}} (\texttt{kws-0007}). ``List every award category at the 59th Daejong Film
Awards with its winner or winning work.'' Key \texttt{[category]}; category exact match, winner/work free-text. 26 rows, 2 columns.
\item \textbf{\textsc{Hard}} (\texttt{kws-mh003}). ``For the 17th--20th presidential elections, list
every candidate with at least 1\% of the vote, with party, vote count, vote share, election date, and
outcome.'' Key \texttt{[election, candidate]} (a 2-D grid); keys exact match, party and outcome free-text, vote count and share numeric tolerance,
date a date tolerance. 15 rows, 7 columns.
\end{itemize}

\subsection{Reproducibility, License, and AI Assistance}
I release the benchmark, the construction pipeline, and the scorer under the MIT license, which covers
the benchmark items and code only, not the linked web content. The tasks were synthesized and verified
by large language models behind automated gates and then audited by a native-speaker spot-check; I
therefore label the benchmark pipeline-verified rather than externally gold-annotated. Because the
scorer parses a model's table directly and grades every cell with one shared, deterministic type-aware
comparator (keyed on each column's declared format), the headline metrics are reproducible exactly from
the released data and code, with no eval-time model call; the released schema additionally ships a
WideSearch-compatible \texttt{eval\_pipeline} that names a per-column metric---and an LLM-judge criterion
for free-text columns---for interoperability with the WideSearch grader.

\subsection{Additional Results}
\paragraph{Category spread.} Pooling Row-F1 by category (Figure~\ref{fig:percategory}, the spread behind
the aggregate), most of the
sixteen categories cluster between 0.30 and 0.40---Sports/Games, the largest at 1{,}600 model-tables, sits
mid-pack at 0.35, alongside Places/Regions (0.39), Entertainment/Media (0.35), Government/Politics (0.30),
and Science/IT (0.30), with History/Culture (0.46) and Economy (0.50) at the top. The clear outlier is
Literature/Books/Language at 0.19 (a small slice, 140 model-tables), where canonical lists are sparse and
attribute values are hardest to pin down on the web. No single category dominates the headline.

\paragraph{Failure by tier.} The two failure modes both track difficulty (Table~\ref{tab:pooled}). Row-F1 falls from 0.44
(\textsc{Easy}) through 0.30 (\textsc{Medium}) to 0.23 (\textsc{Hard}), and parse failures rise in step:
the share of trajectories that return a scorable table drops from 90\% on \textsc{Easy} to about 85\%
on \textsc{Medium} and \textsc{Hard}: the harder tiers more often defeat an agent's ability to emit
the wide 2-D table at all, not only to fill it correctly. I report these as pipeline-verified
diagnostics; a formal multi-annotator agreement study is left to future work.

\subsection{Qualitative Case Studies}
\begin{figure*}[t]
\centering
\includegraphics[width=\textwidth]{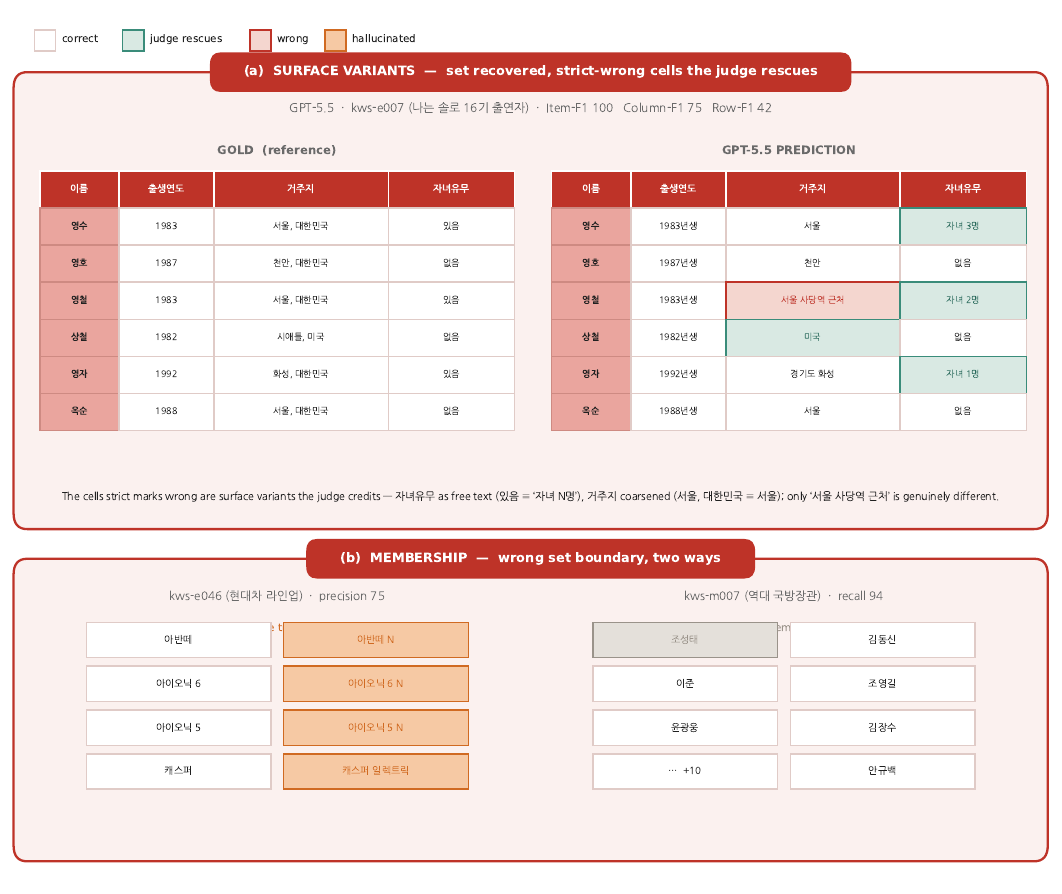}
\caption{\textbf{GPT-5.5 --- surface variants and membership.} Gold vs.\ prediction; predicted cells are
coloured white (correct), teal (strict-wrong but rescued by the semantic judge), red (genuinely wrong),
orange (hallucinated). (a) The cast is fully recovered, and the cells the deterministic scorer marks wrong
are mostly surface variants---the has-children enum written as a free-text count
(``yes'' $\equiv$ ``$N$ children'') and the residence coarsened (``Seoul, Korea'' $\equiv$ ``Seoul'')---that the
semantic judge credits (Table~\ref{tab:judged}); only ``near Sadang Station, Seoul'' is a genuinely different
location. This is the kind of strict-wrong cell that motivates the judged re-scoring. (b) The set boundary
is nearly right both ways, with a few invented product trims and one dropped boundary member.}
\label{fig:case_gpt55}
\end{figure*}

\begin{figure*}[t]
\centering
\includegraphics[width=\textwidth]{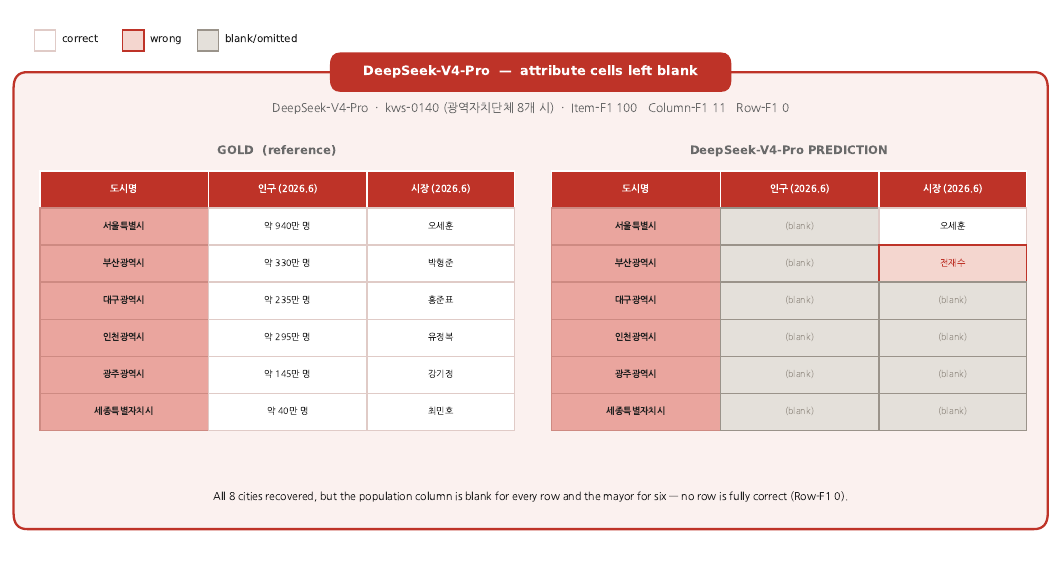}
\caption{\textbf{DeepSeek-V4-Pro --- blank attribute cells.} All eight metropolitan cities are recovered,
but the population column is left blank for every row and the mayor for most, so no row is complete.}
\label{fig:case_deepseek}
\end{figure*}

\begin{figure*}[t]
\centering
\includegraphics[width=\textwidth]{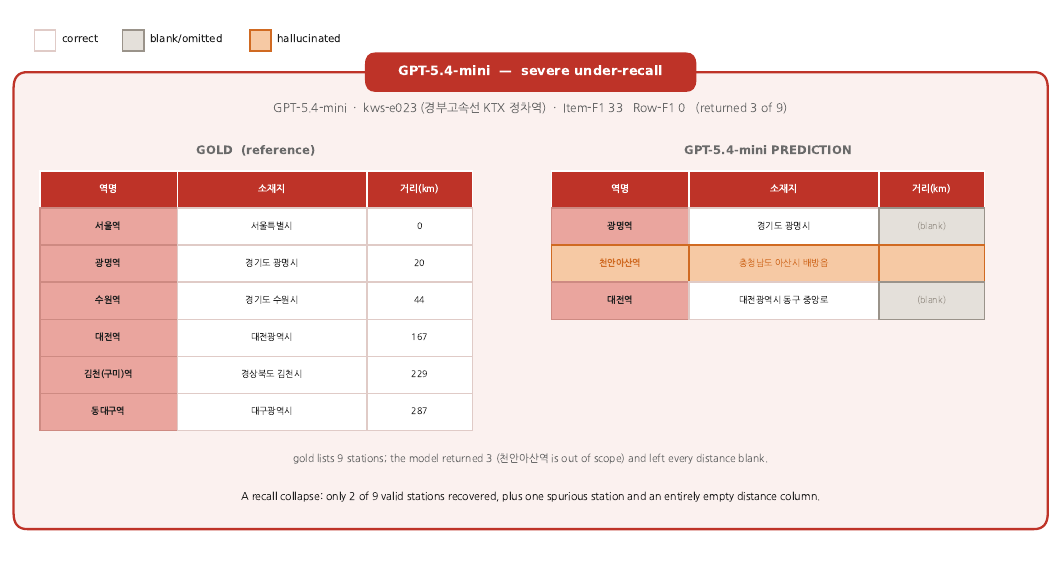}
\caption{\textbf{GPT-5.4-mini --- under-recall.} Of nine valid stations the model returns only three (one
of them out of scope) and leaves every distance blank --- a recall collapse.}
\label{fig:case_gpt54mini}
\end{figure*}

\begin{figure*}[t]
\centering
\includegraphics[width=\textwidth]{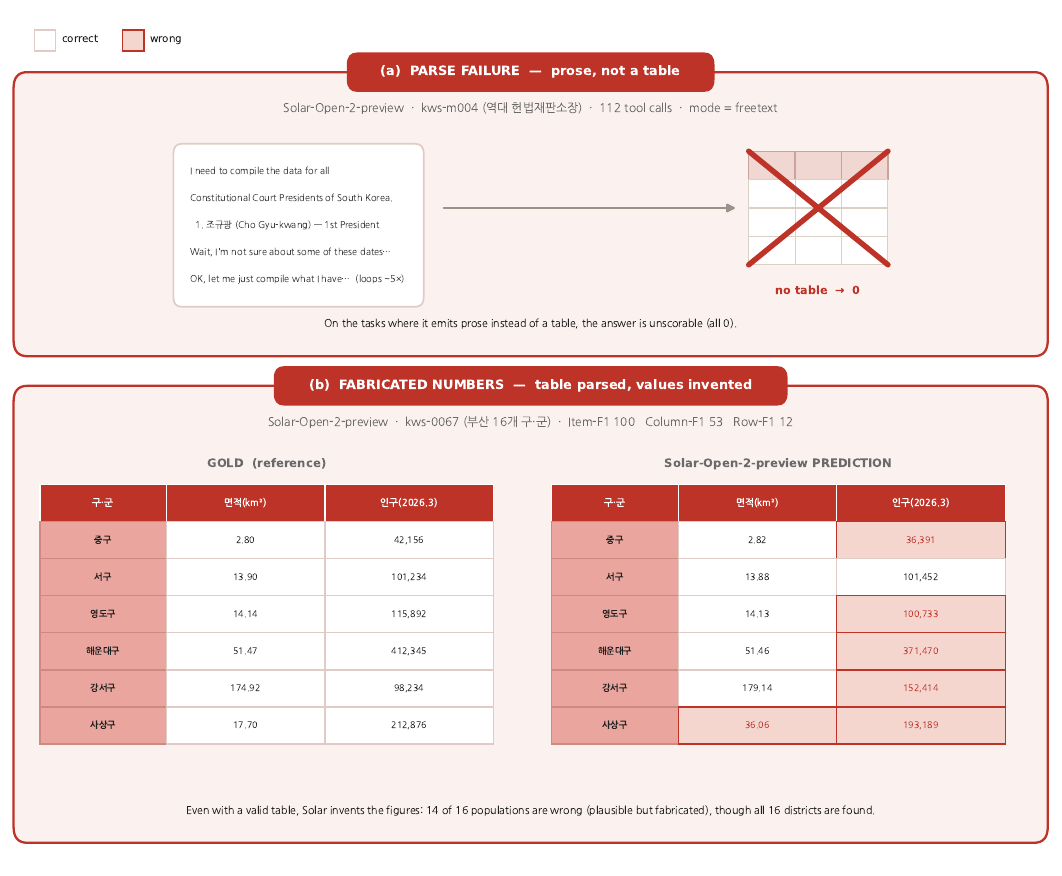}
\caption{\textbf{Solar-Open-2-preview --- two failure modes.} (a) Parse failure: the model narrates a
prose list instead of a table, so the answer is unscorable. (b) Even when it emits a valid table, the
numeric cells are fabricated (plausible but wrong populations), so no row is complete.}
\label{fig:case_solar}
\end{figure*}

\begin{figure*}[t]
\centering
\includegraphics[width=\textwidth]{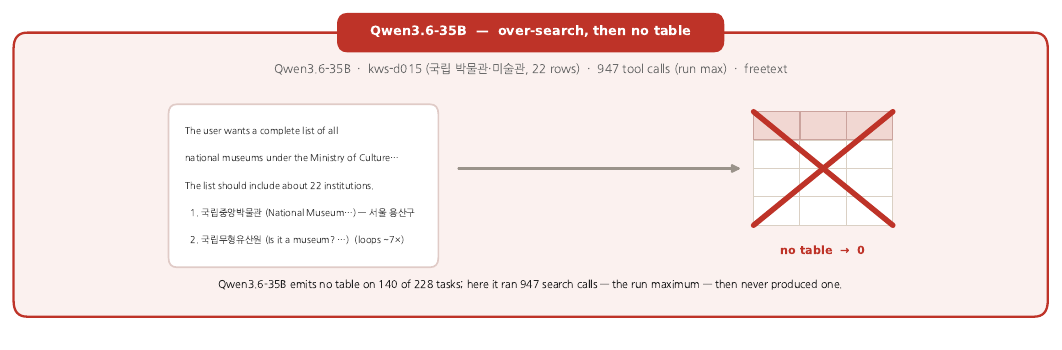}
\caption{\textbf{Qwen3.6-35B --- over-search, then no table.} The model issues the run's maximum number
of tool calls, yet still emits prose rather than a table --- its most common outcome on the benchmark.}
\label{fig:case_qwen}
\end{figure*}

Each model has a characteristic way of failing. I show one real example per model
(Figures~\ref{fig:case_gpt55}--\ref{fig:case_qwen}), every cell verbatim from the released eval records,
with predicted cells colour-coded against gold. Together they span the failure taxonomy behind the
Item$\gg$Row gap: recovering the members but missing the cells takes several distinct forms.

\begin{figure*}[t]
\centering
\includegraphics[width=\textwidth]{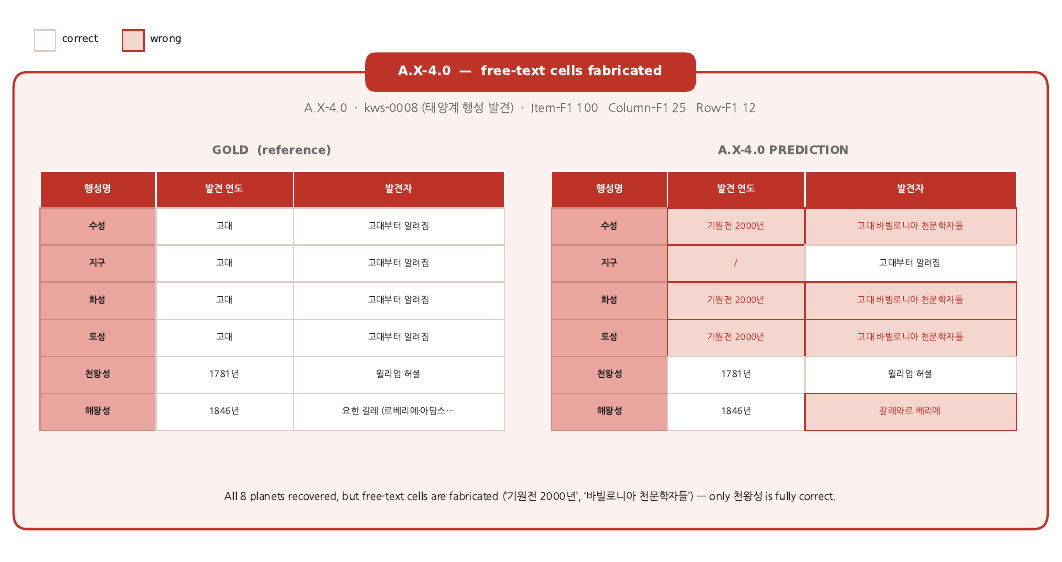}
\caption{\textbf{A.X-4.0 --- fabricated free-text cells.} All eight planets are recovered, but the
discovery-year and discoverer cells are fabricated (``2000 BC'', ``Babylonian astronomers'' where the gold
is ``ancient''); only one row is fully correct.}
\label{fig:case_ax}
\end{figure*}

GPT-5.5 (Figure~\ref{fig:case_gpt55}) recovers the full cast of the variety show \textit{I am SOLO}
season 16 (Item-F1 100) but writes the has-children enum as a free-text count (``3 children'' for the
allowed value ``yes'') and coarsens several addresses (``Seoul'' for ``Seoul, Korea''), so seven of twelve rows break and Row-F1 is 42; its
set boundary is near-perfect either way, with precision 75 from four invented product trims and recall 94
from one dropped boundary member. DeepSeek-V4-Pro (Figure~\ref{fig:case_deepseek}) recovers all eight
metropolitan cities but leaves the population column blank for every row and the mayor for six, so no row
is complete (Row-F1 0). GPT-5.4-mini (Figure~\ref{fig:case_gpt54mini}) collapses on recall, returning only
two of nine valid high-speed-rail stations plus one out-of-scope station, with the distance column left
empty. A.X-4.0 (Figure~\ref{fig:case_ax}) recovers all eight planets but fabricates their free-text
discovery cells (``2000 BC'' and ``Babylonian astronomers'' where the gold is ``ancient''), leaving only
one row correct. The two lowest-scoring self-hosted models often fail before scoring even begins.
Solar-Open-2-preview (Figure~\ref{fig:case_solar}) fails two ways: on many tasks it narrates a prose list
instead of a table (unscorable), and even when it does emit a valid table it fabricates the numbers---14
of 16 district populations wrong, though all 16 are found. Qwen3.6-35B (Figure~\ref{fig:case_qwen}) runs
947 tool calls, the run maximum, yet still emits no table, failing to produce one on 140 of 228 tasks.
For Qwen, structured output, not search, is the binding constraint.

\clearpage

\begin{thebibliography}{}

\bibitem[Amouyal et al. 2023]{amouyal2023qampari} Amouyal, S. J.; Wolfson, T.; Rubin, O.; Yoran, O.; Herzig, J.; and Berant, J. 2023. QAMPARI: An Open-domain Question Answering Benchmark for Questions with Many Answers from Multiple Paragraphs. arXiv:2205.12665.
\bibitem[Asai et al. 2024]{asai2024selfrag} Asai, A.; Wu, Z.; Wang, Y.; Sil, A.; and Hajishirzi, H. 2024. Self-RAG: Learning to Retrieve, Generate, and Critique through Self-Reflection. ICLR.
\bibitem[CAIS and Scale AI 2025]{phan2025hle} CAIS; and Scale AI. 2025. Humanity's Last Exam. arXiv:2501.14249.
\bibitem[Chen et al. 2020a]{chen2020hybridqa} Chen, W.; Zha, H.; Chen, Z.; Xiong, W.; Wang, H.; and Wang, W. Y. 2020. HybridQA: A Dataset of Multi-Hop Question Answering over Tabular and Textual Data. EMNLP Findings.
\bibitem[Chen et al. 2020b]{chen2020tabfact} Chen, W.; Wang, H.; Chen, J.; Zhang, Y.; Wang, H.; Li, S.; Zhou, X.; and Wang, W. Y. 2020. TabFact: A Large-scale Dataset for Table-based Fact Verification. ICLR.
\bibitem[DeepSeek-AI 2024]{deepseek2024v3} DeepSeek-AI. 2024. DeepSeek-V3 Technical Report. arXiv:2412.19437.
\bibitem[Deng et al. 2023]{deng2023mind2web} Deng, X.; Gu, Y.; Zheng, B.; Chen, S.; Stevens, S.; Wang, B.; Sun, H.; and Su, Y. 2023. Mind2Web: Towards a Generalist Agent for the Web. NeurIPS.
\bibitem[Ding et al. 2023]{ding2023ultrachat} Ding, N.; Chen, Y.; Xu, B.; Qin, Y.; Zheng, Z.; Hu, S.; Liu, Z.; Sun, M.; and Zhou, B. 2023. Enhancing Chat Language Models by Scaling High-quality Instructional Conversations. EMNLP.
\bibitem[Dong et al. 2024]{dong2024generalization} Dong, Y.; Jiang, X.; Liu, H.; Jin, Z.; Gu, B.; Yang, M.; and Li, G. 2024. Generalization or Memorization: Data Contamination and Trustworthy Evaluation for LLMs. ACL Findings.
\bibitem[Gemini Team 2023]{gemini2023} Gemini Team. 2023. Gemini: A Family of Highly Capable Multimodal Models. arXiv:2312.11805.
\bibitem[Golchin and Surdeanu 2024]{golchin2024timetravel} Golchin, S.; and Surdeanu, M. 2024. Time Travel in LLMs: Tracing Data Contamination in Large Language Models. ICLR.
\bibitem[He et al. 2024]{he2024webvoyager} He, H.; Yao, W.; Ma, K.; Yu, W.; Dai, Y.; Zhang, H.; Lan, Z.; and Yu, D. 2024. WebVoyager: Building an End-to-End Web Agent with Large Multimodal Models. ACL.
\bibitem[Ho et al. 2020]{ho2020twowiki} Ho, X.; Duong Nguyen, A.-K.; Sugawara, S.; and Aizawa, A. 2020. Constructing A Multi-hop QA Dataset for Comprehensive Evaluation of Reasoning Steps. COLING.
\bibitem[HyperCLOVA X Team 2024]{yoo2024hyperclovax} HyperCLOVA X Team. 2024. HyperCLOVA X Technical Report. arXiv:2404.01954.
\bibitem[Jang et al. 2022]{jang2022kobest} Jang, M.; Kim, D.; Kwon, D. S.; and Davis, E. 2022. KoBEST: Korean Balanced Evaluation of Significant Tasks. COLING.
\bibitem[Jin et al. 2024]{jin2024kobbq} Jin, J.; Kim, J.; Lee, N.; Yoo, H.; Oh, A.; and Lee, H. 2024. KoBBQ: Korean Bias Benchmark for Question Answering. TACL.
\bibitem[Joshi et al. 2017]{joshi2017triviaqa} Joshi, M.; Choi, E.; Weld, D.; and Zettlemoyer, L. 2017. TriviaQA: A Large Scale Distantly Supervised Challenge Dataset for Reading Comprehension. ACL.
\bibitem[Kanana Team 2025]{kakao2025kanana} Kanana LLM Team. 2025. Kanana: Compute-efficient Bilingual Language Models. arXiv:2502.18934.
\bibitem[Kim et al. 2021]{kim2021hyperclova} Kim, B.; Kim, H.; Lee, S.-W.; Lee, G.; Kwak, D.; Jeon, D. H.; Park, S.; Kim, S.; Kim, S.; Seo, D.; Lee, H.; Jeong, M.; Lee, S.; Kim, M.; Ko, S. H.; Kim, S.; Park, T.; Kim, J.; Kang, S.; Ryu, N.-H.; Yoo, K. M.; Chang, M.; Suh, S.; In, S.; Park, J.; Kim, K.; Kim, H.; Jeong, J.; Yeo, Y. G.; Ham, D.; Park, D.; Lee, M. Y.; Kang, J.; Kang, I.; Ha, J.-W.; Park, W.; and Sung, N. 2021. What Changes Can Large-scale Language Models Bring? Intensive Study on HyperCLOVA. EMNLP.
\bibitem[Kim et al. 2024a]{kim2024click} Kim, E.; Suk, J.; Oh, P.; Yoo, H.; Thorne, J.; and Oh, A. 2024. CLIcK: A Benchmark Dataset of Cultural and Linguistic Intelligence in Korean. LREC-COLING.
\bibitem[Kim et al. 2024b]{kim2024solar} Kim, S.; Kim, D.; Park, C.; Lee, W.; Song, W.; Kim, Y.; Kim, H.; Kim, Y.; Lee, H.; Kim, J.; Ahn, C.; Yang, S.; Lee, S.; Park, H.; Gim, G.; Cha, M.; Lee, H.; and Kim, S. 2024. SOLAR 10.7B: Scaling Large Language Models with Simple yet Effective Depth Up-Scaling. NAACL.
\bibitem[Krishna et al. 2024]{krishna2024frames} Krishna, S.; Krishna, K.; Mohananey, A.; Schwarcz, S.; Stambler, A.; Upadhyay, S.; and Faruqui, M. 2024. FRAMES: Factuality, Retrieval, And reasoning MEasurement Set. arXiv:2409.12941.
\bibitem[Kwiatkowski et al. 2019]{kwiatkowski2019nq} Kwiatkowski, T.; Palomaki, J.; Redfield, O.; Collins, M.; Parikh, A.; Alberti, C.; Epstein, D.; Polosukhin, I.; Devlin, J.; Lee, K.; Toutanova, K.; Jones, L.; Kelcey, M.; Chang, M.-W.; Dai, A. M.; Uszkoreit, J.; Le, Q.; and Petrov, S. 2019. Natural Questions: A Benchmark for Question Answering Research. TACL.
\bibitem[Lee et al. 2026]{lee2026kbrowsecomp} Lee, N.; Yoon, D.; Son, G.; Kim, G.; Ko, D.; Park, J.; Yoo, H.; Cho, J.; Park, J.; Lee, C.; Jang, K.; Kim, J.; Kim, E.; Cho, W.; and Kim, S. 2026. K-BrowseComp: A Web Browsing Agent Benchmark Grounded in Korean Contexts. arXiv:2606.02404.
\bibitem[Lewis et al. 2020]{lewis2020rag} Lewis, P.; Perez, E.; Piktus, A.; Petroni, F.; Karpukhin, V.; Goyal, N.; K\"{u}ttler, H.; Lewis, M.; Yih, W.-t.; Rockt\"{a}schel, T.; Riedel, S.; and Kiela, D. 2020. Retrieval-Augmented Generation for Knowledge-Intensive NLP Tasks. NeurIPS.
\bibitem[LG AI Research 2024]{lgai2024exaone} LG AI Research. 2024. EXAONE 3.0 7.8B Instruction Tuned Language Model. arXiv:2408.03541.
\bibitem[Li et al. 2025]{li2025searcho1} Li, X.; Dong, G.; Jin, J.; Zhang, Y.; Zhou, Y.; Zhu, Y.; Zhang, P.; and Dou, Z. 2025. Search-o1: Agentic Search-Enhanced Large Reasoning Models. arXiv:2501.05366.
\bibitem[Lim et al. 2019]{lim2019korquad} Lim, S.; Kim, M.; and Lee, J. 2019. KorQuAD1.0: Korean QA Dataset for Machine Reading Comprehension. arXiv:1909.07005.
\bibitem[Liu et al. 2023]{liu2023geval} Liu, Y.; Iter, D.; Xu, Y.; Wang, S.; Xu, R.; and Zhu, C. 2023. G-Eval: NLG Evaluation using GPT-4 with Better Human Alignment. EMNLP.
\bibitem[Liu et al. 2024]{liu2024agentbench} Liu, X.; Yu, H.; Zhang, H.; Xu, Y.; Lei, X.; Lai, H.; Gu, Y.; Ding, H.; Men, K.; Yang, K.; Zhang, S.; Deng, X.; Zeng, A.; Du, Z.; Zhang, C.; Shen, S.; Zhang, T.; Su, Y.; Sun, H.; Huang, M.; Dong, Y.; and Tang, J. 2024. AgentBench: Evaluating LLMs as Agents. ICLR.
\bibitem[Llama Team 2024]{grattafiori2024llama3} Llama Team. 2024. The Llama 3 Herd of Models. arXiv:2407.21783.
\bibitem[Mialon et al. 2023]{mialon2023gaia} Mialon, G.; Fourrier, C.; Swift, C.; Wolf, T.; LeCun, Y.; and Scialom, T. 2023. GAIA: A Benchmark for General AI Assistants. arXiv:2311.12983.
\bibitem[Mukherjee et al. 2023]{mukherjee2023orca} Mukherjee, S.; Mitra, A.; Jawahar, G.; Agarwal, S.; Palangi, H.; and Awadallah, A. 2023. Orca: Progressive Learning from Complex Explanation Traces of GPT-4. arXiv:2306.02707.
\bibitem[Nakano et al. 2021]{nakano2021webgpt} Nakano, R.; Hilton, J.; Balaji, S.; Wu, J.; Ouyang, L.; Kim, C.; Hesse, C.; Jain, S.; Kosaraju, V.; Saunders, W.; Jiang, X.; Cobbe, K.; Eloundou, T.; Krueger, G.; Button, K.; Knight, M.; Chess, B.; and Schulman, J. 2021. WebGPT: Browser-assisted Question-answering with Human Feedback. arXiv:2112.09332.
\bibitem[Nan et al. 2022]{nan2022fetaqa} Nan, L.; Hsieh, C.; Mao, Z.; Lin, X. V.; Verma, N.; Zhang, R.; Kry\'{s}ci\'{n}ski, W.; Schoelkopf, H.; Kong, R.; Tang, X.; Mutuma, M.; Rosand, B.; Trindade, I.; Bandaru, R.; Cunningham, J.; Xiong, C.; and Radev, D. 2022. FeTaQA: Free-form Table Question Answering. TACL.
\bibitem[OpenAI 2023]{openai2023gpt4} OpenAI. 2023. GPT-4 Technical Report. arXiv:2303.08774.
\bibitem[Oren et al. 2024]{oren2024proving} Oren, Y.; Meister, N.; Chatterji, N.; Ladhak, F.; and Hashimoto, T. B. 2024. Proving Test Set Contamination in Black Box Language Models. ICLR.
\bibitem[Park et al. 2021]{park2021klue} Park, S.; Moon, J.; Kim, S.; Cho, W. I.; Han, J.; Park, J.; Song, C.; Kim, J.; Song, Y.; Oh, T.; Lee, J.; Oh, J.; Lyu, S.; Jeong, Y.; Lee, I.; Seo, S.; Lee, D.; Kim, H.; Lee, M.; Jang, S.; Do, S.; Kim, S.; Lim, K.; Lee, J.; Park, K.; Shin, J.; Kim, S.; Park, L.; Oh, A.; Ha, J.-W.; and Cho, K. 2021. KLUE: Korean Language Understanding Evaluation. arXiv:2105.09680.
\bibitem[Pasupat and Liang 2015]{pasupat2015wtq} Pasupat, P.; and Liang, P. 2015. Compositional Semantic Parsing on Semi-Structured Tables. ACL.
\bibitem[Patil et al. 2024]{patil2024gorilla} Patil, S. G.; Zhang, T.; Wang, X.; and Gonzalez, J. E. 2024. Gorilla: Large Language Model Connected with Massive APIs. NeurIPS.
\bibitem[Qin et al. 2024]{qin2024toolllm} Qin, Y.; Liang, S.; Ye, Y.; Zhu, K.; Yan, L.; Lu, Y.; Lin, Y.; Cong, X.; Tang, X.; Qian, B.; Zhao, S.; Hong, L.; Tian, R.; Xie, R.; Zhou, J.; Gerstein, M.; Li, D.; Liu, Z.; and Sun, M. 2024. ToolLLM: Facilitating Large Language Models to Master 16000+ Real-world APIs. ICLR.
\bibitem[Qwen Team 2024]{qwen2024qwen25} Qwen Team. 2024. Qwen2.5 Technical Report. arXiv:2412.15115.
\bibitem[Romanou et al. 2025]{romanou2025include} Romanou, A.; Foroutan, N.; Sotnikova, A.; Chen, Z.; Nelaturu, S. H.; Singh, S.; Maheshwary, R.; Altomare, M.; Haggag, M. A.; A, S.; Amayuelas, A.; Amirudin, A. H.; Aryabumi, V.; Boiko, D.; Chang, M.; Chim, J.; Cohen, G.; Dalmia, A. K.; Diress, A.; Duwal, S.; Dzenhaliou, D.; Erazo Florez, D. F.; Farestam, F.; Imperial, J. M.; Islam, S. B.; Isotalo, P.; Jabbarishiviari, M.; Karlsson, B. F.; Khalilov, E.; Klamm, C.; Koto, F.; Krzemi\'{n}ski, D.; de Melo, G. A.; Montariol, S.; Nan, Y.; Niklaus, J.; Novikova, J.; Obando Ceron, J. S.; Paul, D.; Ploeger, E.; Purbey, J.; Rajwal, S.; Ravi, S. S.; Rydell, S.; Santhosh, R.; Sharma, D.; Prifti Skenduli, M.; Soltani Moakhar, A.; Soltani Moakhar, B.; Tamir, R.; Tarun, A. K.; Wasi, A. T.; Weerasinghe, T. O.; Yilmaz, S.; Zhang, M.; Schlag, I.; Fadaee, M.; Hooker, S.; and Bosselut, A. 2025. INCLUDE: Evaluating Multilingual Language Understanding with Regional Knowledge. ICLR.
\bibitem[Sainz et al. 2023]{sainz2023contamination} Sainz, O.; Campos, J.; Garc\'{i}a-Ferrero, I.; Etxaniz, J.; Lopez de Lacalle, O.; and Agirre, E. 2023. NLP Evaluation in Trouble: On the Need to Measure LLM Data Contamination for each Benchmark. EMNLP Findings.
\bibitem[Schick et al. 2023]{schick2023toolformer} Schick, T.; Dwivedi-Yu, J.; Dess\`{i}, R.; Raileanu, R.; Lomeli, M.; Zettlemoyer, L.; Cancedda, N.; and Scialom, T. 2023. Toolformer: Language Models Can Teach Themselves to Use Tools. NeurIPS.
\bibitem[Son et al. 2024]{son2024haerae} Son, G.; Lee, H.; Kim, S.; Kim, H.; Lee, J. C.; Yeom, J. W.; Jung, J.; Kim, J. W.; and Kim, S. 2024. HAE-RAE Bench: Evaluation of Korean Knowledge in Language Models. LREC-COLING.
\bibitem[Son et al. 2025]{son2025kmmlu} Son, G.; Lee, H.; Kim, S.; Kim, S.; Muennighoff, N.; Choi, T.; Park, C.; Yoo, K. M.; and Biderman, S. 2025. KMMLU: Measuring Massive Multitask Language Understanding in Korean. NAACL.
\bibitem[Spiesberger et al. 2026]{spiesberger2026softcontamination} Spiesberger, A.; Vazquez, J. J.; Pochinkov, N.; Gaven\v{c}iak, T.; Grietzer, P.; Leech, G.; and Schoots, N. 2026. Soft Contamination Means Benchmarks Test Shallow Generalization. arXiv:2602.12413.
\bibitem[Trivedi et al. 2022]{trivedi2022musique} Trivedi, H.; Balasubramanian, N.; Khot, T.; and Sabharwal, A. 2022. MuSiQue: Multihop Questions via Single-hop Question Composition. TACL.
\bibitem[Vu et al. 2024]{vu2024freshllms} Vu, T.; Iyyer, M.; Wang, X.; Constant, N.; Wei, J.; Wei, J.; Tar, C.; Sung, Y.-H.; Zhou, D.; Le, Q.; and Luong, T. 2024. FreshLLMs: Refreshing Large Language Models with Search Engine Augmentation. ACL Findings.
\bibitem[Wang et al. 2023]{wang2023selfinstruct} Wang, Y.; Kordi, Y.; Mishra, S.; Liu, A.; Smith, N. A.; Khashabi, D.; and Hajishirzi, H. 2023. Self-Instruct: Aligning Language Models with Self-Generated Instructions. ACL.
\bibitem[Wei et al. 2024]{wei2024simpleqa} Wei, J.; Karina, N.; Chung, H. W.; Jiao, Y. J.; Papay, S.; Glaese, A.; Schulman, J.; and Fedus, W. 2024. Measuring Short-form Factuality in Large Language Models. arXiv:2411.04368.
\bibitem[Wei et al. 2025]{wei2025browsecomp} Wei, J.; Sun, Z.; Papay, S.; McKinney, S.; Han, J.; Fulford, I.; Chung, H. W.; Passos, A. T.; Fedus, W.; and Glaese, A. 2025. BrowseComp: A Simple Yet Challenging Benchmark for Browsing Agents. arXiv:2504.12516.
\bibitem[Whitehouse et al. 2026]{whitehouse2026menlo} Whitehouse, C.; Ruder, S.; Lin, T.; Kurylo, O.; Takagi, H.; Lam, J.; Busetto, N.; Diaz, D.; and Guzm\'{a}n, F. 2026. MENLO: Evaluating Native-like Quality Across 47 Languages. arXiv:2509.26601.
\bibitem[Wong et al. 2025]{widesearch2025} Wong, R.; Wang, J.; Zhao, J.; Chen, L.; Gao, Y.; Zhang, L.; Zhou, X.; Wang, Z.; Xiang, K.; Zhang, G.; Huang, W.; Wang, Y.; and Wang, K. 2025. WideSearch: Benchmarking Agentic Broad Info-Seeking. arXiv:2508.07999.
\bibitem[Xu et al. 2024a]{xu2024magpie} Xu, Z.; Jiang, F.; Niu, L.; Deng, Y.; Poovendran, R.; Choi, Y.; and Lin, B. Y. 2024. Magpie: Alignment Data Synthesis from Scratch by Prompting Aligned LLMs with Nothing. arXiv:2406.08464.
\bibitem[Xu et al. 2024b]{xu2024wizardlm} Xu, C.; Sun, Q.; Zheng, K.; Geng, X.; Zhao, P.; Feng, J.; Tao, C.; Lin, Q.; and Jiang, D. 2024. WizardLM: Empowering Large Pre-trained Language Models to Follow Complex Instructions. ICLR.
\bibitem[Yang et al. 2018]{yang2018hotpotqa} Yang, Z.; Qi, P.; Zhang, S.; Bengio, Y.; Cohen, W.; Salakhutdinov, R.; and Manning, C. D. 2018. HotpotQA: A Dataset for Diverse, Explainable Multi-hop Question Answering. EMNLP.
\bibitem[Yao et al. 2022]{yao2022webshop} Yao, S.; Chen, H.; Yang, J.; and Narasimhan, K. 2022. WebShop: Towards Scalable Real-World Web Interaction with Grounded Language Agents. NeurIPS.
\bibitem[Yao et al. 2023]{yao2023react} Yao, S.; Zhao, J.; Yu, D.; Du, N.; Shafran, I.; Narasimhan, K.; and Cao, Y. 2023. ReAct: Synergizing Reasoning and Acting in Language Models. ICLR.
\bibitem[Yoran et al. 2024]{yoran2024assistantbench} Yoran, O.; Amouyal, S. J.; Malaviya, C.; Bogin, B.; Press, O.; and Berant, J. 2024. AssistantBench: Can Web Agents Solve Realistic and Time-Consuming Tasks? EMNLP.
\bibitem[Zheng et al. 2023]{zheng2023mtbench} Zheng, L.; Chiang, W.-L.; Sheng, Y.; Zhuang, S.; Wu, Z.; Zhuang, Y.; Lin, Z.; Li, Z.; Li, D.; Xing, E. P.; Zhang, H.; Gonzalez, J. E.; and Stoica, I. 2023. Judging LLM-as-a-Judge with MT-Bench and Chatbot Arena. NeurIPS.
\bibitem[Zhou et al. 2024]{zhou2024webarena} Zhou, S.; Xu, F. F.; Zhu, H.; Zhou, X.; Lo, R.; Sridhar, A.; Cheng, X.; Ou, T.; Bisk, Y.; Fried, D.; Alon, U.; and Neubig, G. 2024. WebArena: A Realistic Web Environment for Building Autonomous Agents. ICLR.
\bibitem[Zhou et al. 2025]{zhou2025browsecompzh} Zhou, P.; Leon, B.; Ying, X.; Zhang, C.; Shao, Y.; Ye, Q.; Chong, D.; Jin, Z.; Xie, C.; Cao, M.; Gu, Y.; Hong, S.; Ren, J.; Chen, J.; Liu, C.; and Hua, Y. 2025. BrowseComp-ZH: Benchmarking Web Browsing Ability of LLMs in Chinese. arXiv:2504.19314.
\bibitem[Zhu et al. 2026]{aggbench2026} Zhu, H.; Xu, Q.; Li, H.; Liu, Y.; Qiu, H.; Chen, J.; and Jin, J. 2026. Aggregation Queries over Unstructured Text: Benchmark and Agentic Method. arXiv:2602.01355.
\end{thebibliography}
\end{document}